%% file: tmi.tex
\documentclass[journal,twoside,web]{ieeecolor}
\usepackage{tmi}
\usepackage{amsmath,amssymb,amsfonts}
\usepackage{graphicx}
\usepackage{textcomp}
\usepackage{framed,multirow}
\usepackage{latexsym}
\usepackage{url}
\usepackage{threeparttable}
\usepackage{gensymb}
\usepackage{subcaption}
\usepackage{booktabs} 
\usepackage{float}
\usepackage{xcolor}
\usepackage[noend]{algpseudocode}
\usepackage{algorithmicx,algorithm}
\usepackage{epsfig}
\usepackage{enumerate}
\usepackage{makecell}
\usepackage{tabularx}
\usepackage{times}
\usepackage{verbatim}
\usepackage{array}
\usepackage{comment}
\usepackage{multicol}
\usepackage{bbding}
\usepackage{longtable}
\usepackage{colortbl}
\usepackage{tabu}
\usepackage{mathrsfs}
\usepackage{bbm}

\definecolor{light-gray}{gray}{0.9}
\definecolor{cvprblue}{rgb}{0.21,0.49,0.74}
\definecolor{citecolor}{RGB}{119,185,0} 
\definecolor{light-gray}{gray}{0.9}
\definecolor{cvprblue}{rgb}{0.21,0.49,0.74}
\definecolor{citecolor}{RGB}{119,185,0} 
\usepackage[pagebackref,breaklinks,colorlinks,citecolor=citecolor]{hyperref}

\def\BibTeX{{\rm B\kern-.05em{\sc i\kern-.025em b}\kern-.08em
    T\kern-.1667em\lower.7ex\hbox{E}\kern-.125emX}}
    
\begin{document}
\title{FunOTTA: On-the-Fly Adaptation on Cross-Domain Fundus Image via Stable Test-time Training}
\author{
Qian Zeng, Le Zhang, Yipeng Liu, \IEEEmembership{Senior Member, IEEE}, Ce Zhu, \IEEEmembership{Fellow, IEEE}, and
Fan Zhang
\thanks{This work is in part supported by the National Key R\&D Program of China (No. 2023YFE0118600, 2024YFE0100700) and the National Natural Science Foundation of China (No. 62371107). (Corresponding author: Fan Zhang.)}
\thanks{Qian Zeng is with the Glasgow College, University of Electronic Science and Technology of China, Chengdu 611731, China.}
\thanks{Le Zhang, Yipeng Liu are with the School of Information and Communication Engineering, University of Electronic Science and Technology of China, Chengdu 611731, China.}
\thanks{Ce Zhu is with the Glasgow College, University of Electronic Science and Technology of China, Chengdu 611731, China, and also with the School of Information and Communication Engineering, University of Electronic Science and Technology of China, Chengdu 611731, China.}
\thanks{Fan Zhang is with the School of Information and Communication Engineering, University of Electronic Science and Technology of China, Chengdu 611731, China. (e-mail: fan.zhang@uestc.edu.cn).}
}

\maketitle

\begin{abstract}
Fundus images are essential for the early screening and detection of eye diseases. While deep learning models using fundus images have significantly advanced the diagnosis of multiple eye diseases, variations in images from different imaging devices and locations (known as domain shifts) pose challenges for deploying pre-trained models in real-world applications. To address this, we propose a novel Fundus On-the-fly Test-Time Adaptation (FunOTTA) framework that effectively generalizes a fundus image diagnosis model to unseen environments, even under strong domain shifts. FunOTTA stands out for its stable adaptation process by performing dynamic disambiguation in the memory bank while minimizing harmful prior knowledge bias. We also introduce a new training objective during adaptation that enables the classifier to incrementally adapt to target patterns with reliable class conditional estimation and consistency regularization. We compare our method with several state-of-the-art test-time adaptation (TTA) pipelines. Experiments on cross-domain fundus image benchmarks across two diseases demonstrate the superiority of the overall framework and individual components under different backbone networks. Code is available at \href{https://github.com/Casperqian/FunOTTA}{https://github.com/Casperqian/FunOTTA}.
\end{abstract}


\section{Introduction}
\label{sec:introduction}
Fundus images are two-dimensional projections that capture detailed views of the fundus, primarily showcasing the central and peripheral retina, optic disc, and macula~\cite{edupuganti2018automatic,li2021applications}. Unlike other eye scan techniques, such as optical coherence tomography (OCT) and angiography, fundus imaging is one of the most viable non-invasive and cost-effective means of examining the retina for diagnosis, making it ideal for large-scale screening~\cite{edupuganti2018automatic}. To date, fundus images have been widely used for a variety of eye diseases, including diabetic retinopathy and glaucoma, both of which can lead to irreversible blindness. Recent advances in deep learning have largely automated eye disease diagnosis using fundus images, accelerating the diagnosis process, reducing assessment costs, and lowering the burden on healthcare systems~\cite{li2021applications,iqbal2022recent}. In fundus image classification tasks, backbone networks such as VGG-Net~\cite{vggnet}, ResNet~\cite{resnet}, and DenseNet~\cite{densenet} are among the most commonly used. Many novel deep learning-based networks such as DRNet~\cite{HASAN2021drnet}, CF-DRNet~\cite{wu2020cfdrnet}, and Triple-DRNet~\cite{jian2023tripledrnet} have also been developed for fundus image classification, demonstrating remarkable performance. However, despite the notable performance in in-distribution environments, where the test data is assumed to be drawn i.i.d. from the same distribution as the training data~\cite{yang2024generalized}, these networks often struggle with domain shift. This is primarily due to imaging variations between training and testing data caused by varying imaging conditions and/or acquisition sites~\cite{che2023towards,xia2024generalizing}. Such a shift poses a significant challenge in deploying off-the-shelf diagnostic models in real-world applications.

Facing these challenges, the research community develops various domain transfer techniques to improve model robustness to handle domain shifts~\cite{liang2024comprehensive,pei2023multi}. Domain adaptation (DA), which aims to transfer knowledge between domains with differing distributions or characteristics, has garnered high interest in medical image analysis~\cite{liang2024comprehensive,pei2023multi,chen2019synergistic}. Prior efforts have applied DA to reduce the impact of domain shift of fundus images for different eye disease diagnoses by bridging the gap between domains~\cite{sun2020gan,zhou2022domain}. However, patient data privacy concerns hinder the practical application of conventional DA, which requires access to both source and target data for adaptation. 
Source-free unsupervised domain adaptation (SFUDA), a special case within DA, also known as test-time domain adaptation, aims to adapt models without access to source data, addressing privacy concerns~\cite{liang2024comprehensive}. However, SFUDA typically requires multiple offline epochs over target data to re-train the model, limiting its feasibility in real-time clinical applications where data arrive sequentially.
Another domain transfer technique is called domain generalization (DG), which operates solely during the training phase to seek domain-invariant representations to enhance model robustness~\cite{liang2024comprehensive,atwany2022drgen,chokuwa2023generalizing}. Popular DG methods used for fundus images focus on image augmentation to stimulate distribution shifts in unseen domains~\cite{che2023towards}. However, most of these approaches rely on multiple source domains to generalize the model to a target domain. This strategy can be challenging to achieve in the medical context, where data is often scarce and hospitals face restrictions on sharing patient data. Test-time adaptation (TTA), stands out for its flexible adaptation setting. Specifically, TTA is highly feasible for clinical practice as it requires only a single-domain-trained model and the target unlabeled data, without relying on the source data like DA~\cite{liang2024comprehensive}.
Additionally, TTA offers the advantage of leveraging target data in an online manner through individual or mini-batch target samples, a process known as online test-time adaptation (OTTA)~\cite{liang2024comprehensive}. This enables immediate response to domain shifts during real-world clinical workflows, distinguishing it from SFUDA methods that rely on offline processing.
Apart from covariate shift scenarios, where domain shifts are primarily driven by changes in the feature, another family of TTA targets label shift, where the label distribution changes at test time. 
However, for common conditions like glaucoma and diabetic retinopathy, we primarily focus on covariate shift.

However, applying the TTA paradigm in a fundus image diagnosis model remains a boundary to explore. Existing TTA methods are primarily tailored for natural images with large inter-class variations~\cite{zhang2019medical}. 
In contrast, fundus images typically exhibit fine-grained differences between classes, which can easily lead to the failure of the existing TTA methods~\cite{tnn2024miccai}.
Another notable challenge is that many TTA methods heavily depend on prior knowledge from the source domain, e.g., the predictive power of the source-trained classifier~\cite{wang2020tent,iwasawa2021t3a}. 
While this reliance can be effective in general computer vision tasks, it does not always transfer well to fundus imaging, where domain shifts can outweigh inter-class differences.  
As a result, even a slight change in the source model may cause a huge decline in performance or result in a degenerate solution.
For this reason, popular TTA operations such as test-time augmentation or entropy minimization may introduce harmful information during adaptation~\cite{wang2020tent,zhang2022memo}. By the same token, the entropy-based memory bank commonly used in OTTA~\cite{iwasawa2021t3a,jang2022tast}, which stores features and logits based on entropy selection for the refinement of pseudo-labels, may also accumulate unreliable features.

In this paper, we propose a novel Fundus On-the-fly Test-Time Adaptation (FunOTTA) framework using stable feature learning for diabetic retinopathy (DR) grading and glaucoma diagnosis using fundus images. 
To maximize the potential of training-based methods while avoiding instability, we introduce four key components in this work:
1) A dynamic filtering mechanism that selectively identifies and retains informative instances for memory bank updates. Unlike previous entropy-based filters, our method focuses on feature disambiguation and minimizing reliance on harmful prior knowledge.
2) A trainable prototypical classifier based on ensemble learners and enhanced with a neighbor search strategy. This enables prototypes to dynamically adapt to the evolving target feature space, significantly improving both robustness and expressiveness.
3) A confidence-guided contrastive loss that constructs prototype-to-feature pairs guided by prediction confidence. This loss encourages separation between uncertain instances and closer prototypes, while diversifying features in the latent space.
4) A dual-level alignment strategy that allows the classifier to be trainable and free from harmful biases learned in the source domain, enhancing the prediction reliability and improving the prototype construction.

The key contributions of this work include the following aspects:
\begin{itemize}
    \item We propose the first training-based TTA framework for fundus image classification, with the potential for broader application to other medical imaging domains. This framework demonstrates the potential of training-based TTA in medical contexts, challenging the assumption in prior literature that training-based TTA degrades performance for fundus images.
    \item We introduce the FunOTTA for fundus image diagnosis, achieving superior performance over several state-of-the-art (SOTA) methods on large-scale fundus image benchmarks.
    \item We highlight that prior knowledge learning in the source domain may introduce harmful bias, explaining why conventional entropy-based methods fail in the medical context. To mitigate this, we introduce a novel mechanism to reduce this reliance on such biased knowledge.
\end{itemize}

\begin{table}[th]
\caption{
Adaptation and generalization settings differ by their data and therefore losses during training and testing.
}
\label{tab:settings}
\begin{center}
\resizebox{\linewidth}{!}{
\setlength{\tabcolsep}{3pt}
\begin{tabular}{@{}lcccc@{}}
\bf setting & \bf source data & \bf target data & \bf train loss & \bf test loss \\
\toprule
fine-tuning & - & $x^t, y^t$ & $\mathcal{L}(x^t, y^t)$ & - \\
domain adaptation & $x^s$, $y^s$ & $x^t$ & $\mathcal{L}(x^s, y^s, x^t)$& - \\
domain generalization & $x^s$, $y^s$ & $x^t$ & $\mathcal{L}(x^s, y^s)$ & - \\
test-time adaptation & - & $x^t$ & - & $\mathcal{L}(x^t)$ \\
\bottomrule
\end{tabular}
}
\end{center}
\end{table}

\begin{figure*}[th]
    \centering
    \includegraphics[width=\linewidth]{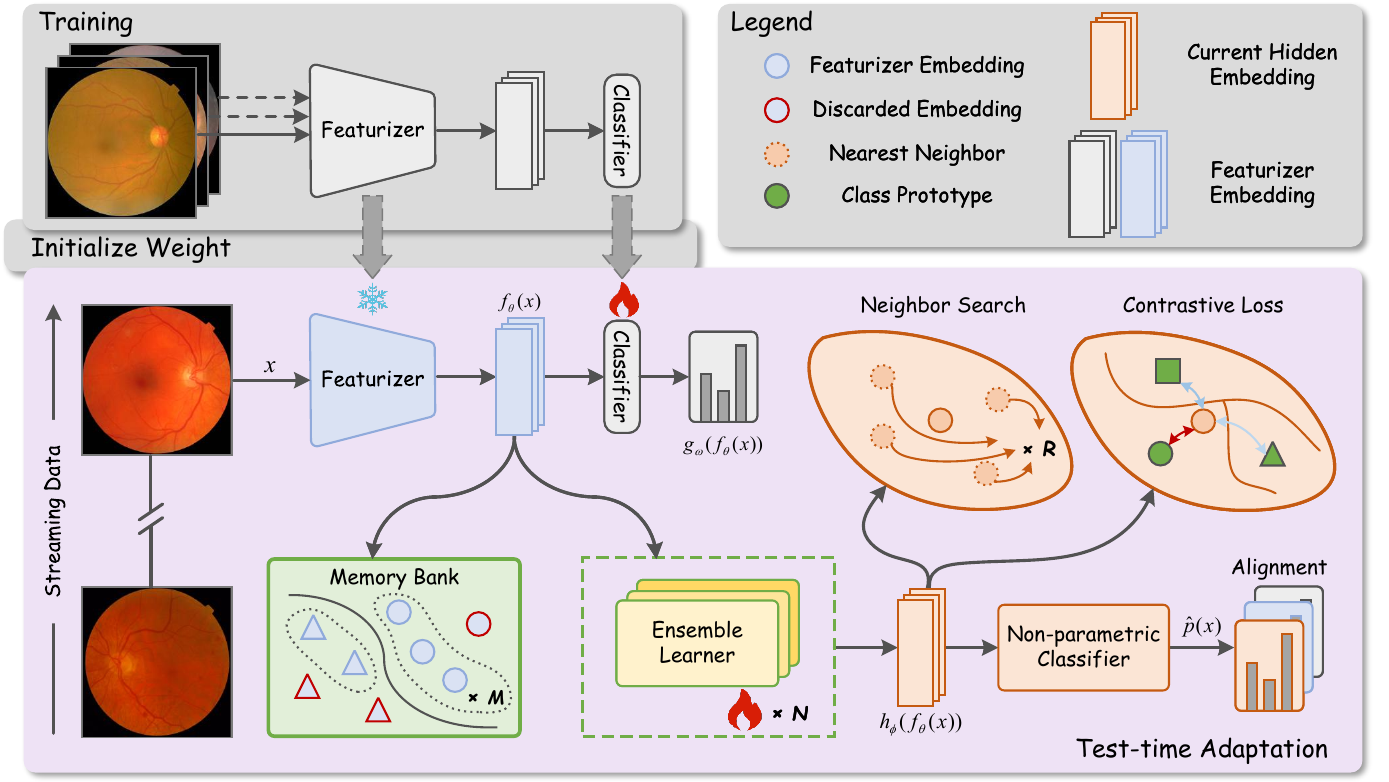}
    \caption{Method overview of FunOTTA to address domain shifts in streaming fundus images. Input images $x$ are fed into the source-trained feature extractor $f_\theta$ to obtain latent features, followed by a memory bank with dynamic filtering mechanism for feature-level disambiguation. Latent features $f_\theta\left(x\right)$ are then processed by an ensemble learner $h_\phi$ to generate low-dimensional embeddings, which are then utilized by a non-parametric classifier to aggregate neighbor information for final predictions. Consistency alignment and contrastive learning paradigm are employed to ensure stable adaptation.}
    \label{fig:overview}
\end{figure*}

\section{Related Works}
\label{sec:related work}
\subsection{Domain Transfer learning}
Domain transfer learning aims to improve model transferability against domain shifts~\cite{liang2024comprehensive,zhou2022dgsurvey}.  A model $f_\theta(x)$ with parameters $\theta$, trained on source data $D^{\text {train }}=\left\{\left(x^s, y^s\right)\right\}$, may struggle to generalize when tested on shifted target data $D^{\text {train }}=\left\{\left(x^s, y^s\right)\right\}$.
Tab.~\ref{tab:settings} outlines the major adaptation and generalization settings, including the required data and the types of losses. 
Existing adaptation and generalization settings often need more access to data and supervision. Fine-tuning, for example, demands both target data and labels to re-train the source model, forcing it to learn new representations under supervision—an approach that is both strict and inefficient. 
While DA uses source data at testing and DG leverages cross-domain data in training, TTA allows for a more restrictive setting, adapting a single-domain trained model without simultaneous access to source, target, or supervision.

{
Recently, a competing paradigm, source free active domain adaptation (SFADA), has emerged as a natural extension of DA, where the source data is inaccessible, yet a limited annotation budget is available in the target domain~\cite{lyu2024sfada}.
This setting introduces a practical trade-off between labeling cost and adaptation performance, making it particularly appealing in real-world scenarios such as medical imaging~\cite{wang2024dual,wang2024advancing,luo2024uncertainty}.
Unlike conventional TTA, which adapts to target distributions in a fully unsupervised manner, SFADA incorporates an active learning process to selectively annotate the most informative samples.
Given their conceptual proximity, many principles underlying TTA can be effectively transferred to the SFADA framework when coupled with an informative instance sampler and an oracle-based querying strategy.
Integrating TTA concepts into the SFADA paradigm therefore represents a promising and scalable direction toward label-efficient and robust domain adaptation, especially in medical imaging scenarios.
}

\subsection{Test Time Adaptation}
Popular TTA methods can be broadly classified into two categories: training-free and training-based TTA. Training-free TTA, without fine-tuning the model’s parameters, aims to adapt the model by a prototype-based classifier~\cite{iwasawa2021t3a,tnn2024miccai,zhong2025omnisam}. In~\cite{iwasawa2021t3a}, Iwasawa et al. obtain pseudo-prototype by the source-trained classifier, and then compute the pseudo-label for target data by measuring the distance to the prototypes. Ambekar et al.~\cite{tnn2024miccai} also use prototypes but apply a non-parametric neighbor search to adjust the final prediction. On the other hand, training-based TTA focuses on generalizing the source model by transforming its source parameters from $\theta_{s}$ to $\theta_{t}$. To enhance the efficiency of TTA, current methods optimize only a small portion of the total parameters, typically the batch normalization layers or the classifier weights. Unsupervised loss functions like entropy minimization~\cite{wang2020tent} or pseudo-labeling~\cite{lee2013pseudo,liang2020shot} are also crucial for improving the alignment between the source and target domains.

Beyond this, a number of broader domain transfer paradigms, including adversarial adaptation, continual learning, and multi-task learning, can also be integrated with TTA.
Adversarial adaptation is one of the most widely used approaches in DA~\cite{zhou2022domain}; however, it typically relies on the availability of multiple domains to explore domain discrepancies, which is often impractical in medical imaging. Continual learning, by contrast, offers a more adaptable framework for incremental domain transfer. In fact, TTA can naturally extend into continual learning settings. For example, CoTTA~\cite{wang2022cotta} has shown strong performance on natural images, suggesting potential for fundus imaging. However, catastrophic forgetting—where the model loses previous knowledge during continuous adaptation—remains a major challenge, especially for medical images, which contain much finer structural and textural details compared to natural images. Multi-task learning (MTL) jointly learns related tasks through shared representations or parameter transfer~\cite{kollias2024distribution}. While effective, modern MTL typically assumes large, multi-task annotated datasets~\cite{kollias2024distribution}, which are costly and difficult to obtain in medical imaging. Moreover, cross-domain MTL approaches require multi-domain data, which is further restricted by privacy concerns~\cite{shen2021variational,long2017learning}. In this work, we focus exclusively on comparing TTA methods, as using multi-task data during training or accessing multiple domains either simultaneously or continuously during inference would violate the assumptions of TTA and introduce inconsistencies in experimental settings, making such comparisons unfair and beyond the scope of this study.

\subsection{Domain Transfer Techniques for Fundus Images}
The data distribution of fundus images across different hospitals or even different acquisition devices can vary significantly, requiring the model to generalize to a distribution different from the source domain it was trained on. To address this challenge, DA methods have emerged and achieved state-of-the-art performance in many fundus image applications.  Among them, adversarial learning-based DA methods have become prevalent in fundus image adaptation, demonstrating outstanding performance in tasks such as fundus image quality assessment~\cite{shen2020domain}, classification~\cite{zhou2022domain}, and segmentation~\cite{wang2019patch}. DG methods are also widely used in fundus image analysis. For instance, Che et al.~\cite{che2023towards} employ task-specific augmentation to stimulate the target domain, while Chokuwa and Khan~\cite{chokuwa2023generalizing} construct a variational autoencoder to seek domain-invariant representations. However, these methods often face performance degradation in real-world applications due to privacy concerns and the scarcity of medical data.

While TTA has shown advanced performance in domain transfer learning, TTA for fundus imaging analysis remains largely unexplored. Among a few TTA methods for fundus image classification, Ambekar et al.~\cite{tnn2024miccai} developed a training-free TTA mechanism that uses an entropy-based filter to construct a memory bank following
~\cite{iwasawa2021t3a}, and then generates the pseudo label for the incoming target data by selected neighbors. Although many training-free TTA methods are relatively less competitive than training-based ones in computer vision tasks, they highlight that most training-based TTA methods fail to adapt the model to the target domain in the medical context. Huang et al.~\cite{huang2023fourier} also point out that fundus image adaptation is challenging, as evidenced by their observation of the upper bound on different fundus image domains (e.g., the fine-tuning performance on the target domain). We conclude three key reasons that training-based TTA methods struggle in fundus images: 
\begin{itemize}
    \item [1)] Fundus imaging datasets available contain only a small fraction of the sample size that the parametric methods are originally designed to handle.
    \item [2)] TTA is originally designed for the computer vision domain, where the semantic gap between classes is typically significant. However, in the context of fundus imaging, inter-class differences are often subtle, making it easier for domain variations to introduce spurious correlations between similar classes.
    \item [3)] Medical diagnosis models are often highly parameter-sensitive. Mainstream TTA operatioins, such as entropy minimization or test-time augmentation, originally devised for natural images, can lead to catastrophic forgetting or degenerate solutions in medical context. 
\end{itemize}

\section{Method}
\label{sec:method}
Fig.~\ref{fig:overview} provides an overview of the FunOTTA framework. In real-world applications, test data may not be all pre-prepared but rather continuously added. Thus, the basic idea of the FunOTTA is to update the weights of the source model using unlabeled target data in an online manner. The trained source model can be decomposed into a linear classifier $g_{\omega}$ and a feature extractor $f_{\theta}$, where $\omega$ and $\theta$ denote the parameters. For efficient adaptation, only the linear classifier is kept trainable, while all other components remain frozen. Details of our approach are explained below.

\subsection{Dynamic Filtering Mechanism}
First, we introduce a dynamic memory bank to store the target data features, denoted by $\mathbb{S}_t$, where $t$ denotes the time step. Following~\cite{iwasawa2021t3a}, at time $t$ = 0, we initialize the memory bank with the L2 norm of the linear classifier’s weights. Given an input sample $x$, at time $t$, $\mathbb{S}_t$ is updated as follows:
\begin{equation}
    \mathbb{S}_t= \mathbb{S}_{t-1} \cup\left\{\frac{f_{\theta}(x)}{\left\|f_{\theta}(x)\right\|}\right\}
\end{equation}
Unlike most strategies that rely on the predictive power of a source-trained classifier (e.g., prediction entropy) to filter the memory bank~\cite{tnn2024miccai,iwasawa2021t3a,jang2022tast,zhang2023unidg}, we dynamically update the memory bank in a more self-supervised manner, alleviating the source-bias and error accumulation issues caused by the distribution shift between training and test data (see our discussion in Sec.~\ref{sec:Managing Entropy for Reliable TTA}). In each time step, ambiguous features are eliminated from the memory based on the distance between class centroids, $\boldsymbol{c}_t = \left\{ c_t^1, c_t^2, \ldots, c_t^K \right\}$, where $K$ is the class number. At the initial time step t = 0, the class centroids are computed as the mean of the features in the memory bank: 
\begin{equation}
    c_t^i = \frac{1}{|\mathbb{S}_t^i|} \sum_{\boldsymbol{z} \in \mathbb{S}_t^i} \boldsymbol{z}
\end{equation}
where $\mathbb{S}_t^i$ denotes the feature set belonging to a specific class at time $t$, and $\boldsymbol{z}$ is the L2-normalized embedding stored in the memory (i.e., $f_{\theta}(x)$). Note that only at $t$ = 0 are the centroids initialized based on the source classifier. For subsequent time steps, our dynamic filtering mechanism updates the centroids by learning a clearer decision boundary to eliminate ambiguous features. 
In particular, our memory bank filters out unreliable information depending on distance in low-level embedding space, as justified by the Johnson–Lindenstrauss (JL) theorem~\cite{jllemma}:
	
    Let $\{x_i \in \mathbb{R}^m : i = 1, \dots, P\}$ be a set of points in a high-dimensional space. The JL lemma ensures that if the target dimension $n$ satisfies $n \geq c \epsilon^{-2} \log P$ for constant $c$ and error tolerance $\epsilon \in (0,1)$, then there exists a projection $A : \mathbb{R}^m \rightarrow \mathbb{R}^n$ such that the pairwise distances between all points are approximately preserved within a multiplicative factor of $(1 \pm \epsilon)$. Specifically, for all $i \ne j$, the following inequality holds:
    \begin{equation}
        (1 - \epsilon) \leq \frac{\|A(x_i) - A(x_j)\|}{\|x_i - x_j\|} \leq (1 + \epsilon).
    \end{equation}
    In our dynamic filtering mechanism, the source-trained featurizer is treated as the projection $A$, mapping inputs into the low-level embedding space. Since this featurizer is trained on the source domain, it preserves the relative distances between inputs in the embedding space to a meaningful degree. These distances serve as a proxy for input relation, as ensured by JL theorem. As a result, our memory bank effectively filters out ambiguous or unreliable inputs—those lying far from the current centroids—by leveraging the geometric structure preserved in the embedding space.
To do so, the objective can be expressed as: 
\begin{equation}
    \underset{\mathbb{S}_t}{\arg \min} \sum_{i=1}^k \sum_{\mathbf{x} \in \mathbb{S}_t^i} \left\| \boldsymbol{z} - \boldsymbol{c}_t^i \right\|^2 = \underset{\mathbb{S}_t}{\arg \min} \sum_{i=1}^k \left| \mathbb{S}_t^i \right| \operatorname{Var} \mathbb{S}_t^i
\end{equation}
where $|\mathbb{S}_t|$ is the size of the memory bank. The centroids are now updated to reflect a more precise decision boundary learned by this objective, without relying on prior knowledge from the source classifier. Once the decision boundary is learned, unreliable features can be filtered out by their distances to the updated centroids:  
\begin{equation}
    \mathbb{S}_t^k=\left\{\boldsymbol{z} \mid \boldsymbol{z} \in \mathbb{S}_t^k, \left\| \boldsymbol{z}- \boldsymbol{c}_t^k \right\| \leq \alpha^k_M\right\}
\end{equation}
where $\alpha^k_M$ is the $M$-smallest L2 distance in $\mathbb{S}_t^k$. This process is repeated iteratively for each incoming batch of data, and the initial centroids for each time step are initialized based on the centroids from the previous time step.

\subsection{Class Conditional Estimation}
To update the model parameters, we estimate the class conditionals of unlabeled target data using the prototypical network, rather than simply relying on the classifier’s output as pseudo labels for self-training. By avoiding the use of the source linear classifier, we further reduce reliance on source-domain bias—analogous to how our method also reduces dependence on prediction entropy.
We first utilize an ensemble learner, denoted as $h_\phi(z)=\frac{1}{N} \sum_{i=1}^N h_{\phi_i}(z)$ to obtain the hidden prototype of each class in the embedding space of $f_\theta \circ h_{\phi i}$, where $h_{\phi i}$ is the $i$-th module, and $N$ is the module number. The prototype can then be represented as:
\begin{equation}
\begin{gathered}
\mu_i^k=\frac{1}{\left|\mathbb{T}^k\right|} \sum_{\boldsymbol{z} \in \mathbb{T}^k} h_{\phi_i}(\boldsymbol{z}) \\
\mathbb{T}^k=\left\{\boldsymbol{z} \in \mathbb{S}_t \mid \arg \max _c g_\omega(\boldsymbol{z})=k\right\}
\end{gathered}
\end{equation}
Using the prototype computed by the non-parametric classifier, we can predict the corresponding class conditionals for all embeddings in the memory bank:
\begin{equation}
    p_i^{\text{proto}}(k \mid z)=\frac{\exp \left(-d\left(h_{\phi_i}(z), \mu_i^k\right) / \tau\right)}{\sum_c \exp \left(-d\left(h_{\phi_i}(z), \mu_i^c\right) / \tau\right)}
\label{eq:class_conditionals}
\end{equation}
where $\tau$ is the temperature coefficient, and $d$ is a distance function. In principle, the distance function can be any distance metric, but we opt for cosine similarity empirically. Following the JL lemma~\cite{jllemma}, similar cases in the higher dimensional image space are positioned close to each other in the source learned lower dimensional embedding space. Hence, we identify the nearest neighbors $\mathcal{N}(x)$ of each input test data within $\mathbb{S}$:
\begin{equation}
    \mathcal{N}(x)=\left\{\boldsymbol{z} \in \mathbb{S}_t \mid d\left(f_\theta(x), \boldsymbol{z}\right) \leq \beta_R\right\}
\end{equation}
where $\beta_{R}$ is the distance between $x$ and the R-closest neighbor. Once identify the neighbors of the input target data, we obtain both the predictions and pseudo-labels aggregated with neighbor information:
\begin{equation}
\begin{gathered}
p_i(k \mid x)=p_i^{\text{proto}}(k \mid f_{\theta}(x)) \\
\widehat{p}_i(k \mid x)=\frac{1}{R} \sum_{z \in \mathcal{N}(x)} \mathbbm{1}\left[\operatorname{argmax}_c p_i^{\text {proto}}(c \mid z)=k\right]
\end{gathered}
\end{equation}
Finally, we construct a straightforward test-time training objective to optimize the ensemble learner, as shown below.
\begin{equation}
    \mathcal{L}_{\mathrm{TTT}}=\frac{1}{N} \sum_{i=1}^N \operatorname{CE}\left(\hat{p}_i(c \mid x), p_i(c \mid x)\right)
\end{equation}

\begin{figure*}[th]
    \centering
    \includegraphics[width=\linewidth]{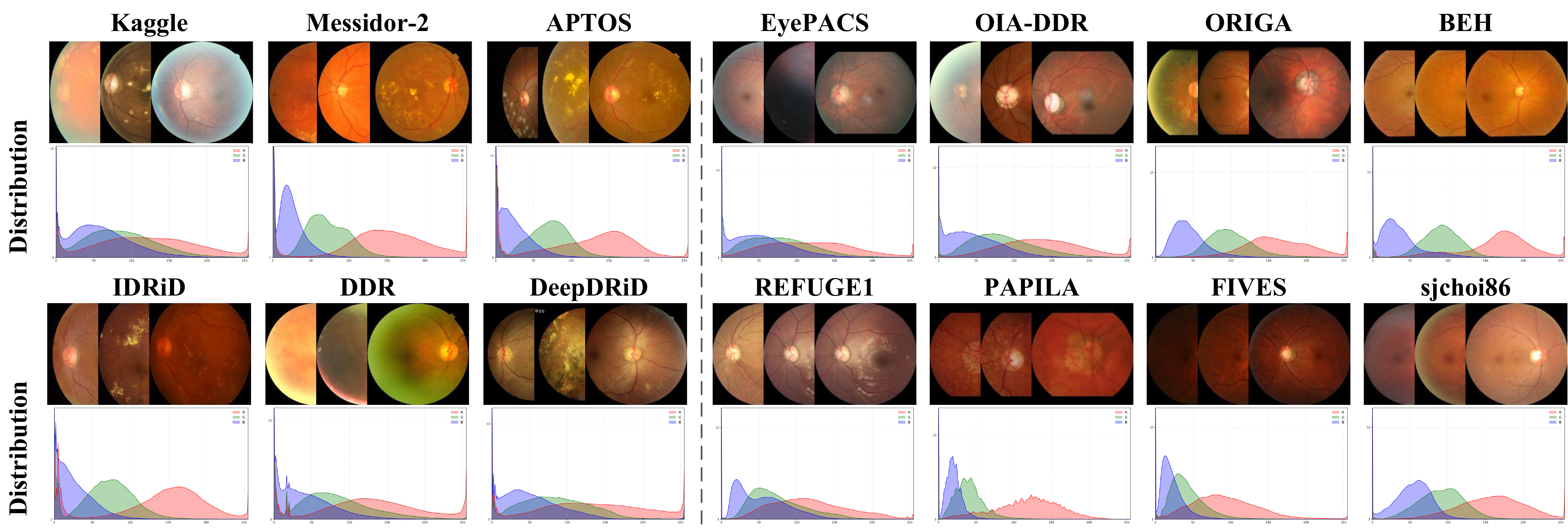}
    \caption{Data sample and RGB statistics of fundus images across different sites.}
    \label{fig:data_sample}
\end{figure*}

\subsection{Confidence-guided Contrastive Loss}
While the aforementioned neighbor search helps generate better pseudo labels, domain shifts—especially when the feature extractor is frozen—can lead to ambiguous representations, making it difficult for simple neighbor search strategies to reliably classify or distinguish target samples.
In recent years, self-supervised contrastive learning has been shown to effectively exploit the pairwise information among target samples, which is beneficial for pseudo-labeling~\cite{chen2022contrastive,bovsnjak2023semppl,xue2023superficial}. Building on this idea, recent works have further explored prototypical contrastive learning to constrain the distance between confident samples and ambiguous ones, thereby improving the recognition of ambiguous instances~\cite{zhou2023learning}.
However, these methods often rely on test-time augmentation to generate positive and negative pairs, which presents challenges in the medical context. As discussed in Section.~\ref{sec:related work}, augmentations commonly used in computer vision may be too aggressive for fundus images. Plus, the time required for these augmentations becomes impractical for OTTA when dealing with streaming data in medical applications.

To address these challenges, we construct positive and negative pairs using ready-made feature embeddings, which reduces the risks associated with augmentation while significantly enhancing efficiency. At each time step, the prototype that shares the same label as the current sample serves as the positive pair, while other prototypes act as negative pairs. There are two main reasons for using prototypes to create pairs: First, prototypes aggregate diverse image information from the memory bank, effectively combining similar image semantics as a unified pair. Second, since the prototype is updated at each time step, these pairs remain dynamic, providing a fresh, feature-level augmentation with each update. This approach functions as an implicit augmentation. However, selecting positive and negative pairs can be challenging, especially in target domains with significant distribution shifts. To improve this process, we introduce a confidence-based restriction that penalizes samples with uncertain predictions, reducing the likelihood of mistakenly associating them with incorrect prototypes. The final confidence-guided contrastive loss for a single module in the ensemble learner is defined as:

\begin{align}
    \mathcal{L}_{CCL} = & -\frac{\exp \left(-H(x)\right)}{\sum_B \exp \left(-H(x)\right)} \cdot \notag \\
              & \log \left( \frac{\exp \left[d\left(h_{\phi}(f_{\theta}(x)), h_{\phi}(\mu_{+})\right)\right]}
              {\sum_{i=0}^k \exp \left[d\left(h_{\phi}(f_{\theta}(x)), h_{\phi}(\mu_i)\right)\right]} \right)
\end{align}

where $B$ is the batch size, and $H(x)=-\sum_{c} p(c \mid x) \cdot \log (p(c \mid x))$. The $L_{CCL}$ is then averaged across the ensemble learner.

\subsection{Dual-level Alignment}
Previous methods create prototypes using the frozen classifier~\cite{tnn2024miccai,iwasawa2021t3a,jang2022tast}; however, this can introduce heavy source bias, which is particularly detrimental in challenging target domains. Additionally, our method integrates confidence information within the contrastive function. If the classifier remains frozen, it may introduce misaligned information due to unreliable predictions. To update the source classifier, we introduce a dual-level alignment loss to adapt it to the unseen target domain stably:
\begin{equation}
\begin{aligned}
    \mathcal{L}_{DAL} = & D_{KL}\left[\hat{p}(x) \| g_{\omega}(f_{\theta}(x))\right] \\
    + & D_{KL}\left[g_{\stackrel{*}{\omega}}(f_{\theta}(x)) \| g_{\omega}(f_{\theta}(x))\right]
\end{aligned}   
\end{equation}

where $D_{KL}$ denotes the Kullback–Leibler (KL) divergence, and $\stackrel{*}{\omega}$ represents the parameters of the frozen classifier.  The second term of the $L_{DAL}$ functions as a consistency loss, helping to prevent catastrophic forgetting in extreme cases.

The final objective is the combination of all the aforementioned losses, which is formulated as:
\begin{equation}
    \mathcal{L} = \mathcal{L}_{TTT}+\lambda_{1} \mathcal{L}_{CCL}+ \lambda_{2} \mathcal{L}_{DAL}
\label{eq:overall_objective}
\end{equation}
where $\lambda_{1}$ and $\lambda_{2}$ are the weights to balance different loss components. 

\begin{table}[th]
    \centering
    \caption{Counts of the source and target domain data. The star symbol ($\star$) denotes the subset of the original dataset.}
    \label{tab:statistics}
    \resizebox{\linewidth}{!}{    
    \begin{tabular}{lcccccc}
    \toprule
    \multirow{2}{*}{DR} & \multicolumn{1}{c}{\textbf{SOURCE}}&\multicolumn{5}{c}{\textbf{TARGET}} \\ 
    \cmidrule(lr){2-2} \cmidrule(lr){3-7}
     & Kaggle$^{\star}$ & Messidor-2 & APTOS & IDRID & OIA-DDR & DeepDRiD \\ 
    \midrule
    Normal       & \multirow{5}{*}{708} & 1017 & 1805 & 168 & 6266 & 540 \\
    Mild         &  & 270 & 370 & 25 & 630 & 140 \\
    Moderate     &  & 347 & 999 & 168 & 4477 & 234 \\
    Severe       &  & 75 & 193 & 93 & 236 & 214 \\
    Proliferate  &  & 35 & 295 & 62 & 913 & 72 \\
    \midrule
    Total        & 3540 & 1744 & 3662 & 516 & 12522 & 1200 \\
    \bottomrule
    \end{tabular}
    }
    \vspace{5pt}
    \\
    \resizebox{\linewidth}{!}{
    \begin{tabular}{lcccccccc}
    \toprule
    \multirow{2}{*}{Glaucoma} & \multicolumn{2}{c}{\textbf{SOURCE}}&\multicolumn{6}{c}{\textbf{TARGET}} \\ 
    \cmidrule(lr){2-3} \cmidrule(lr){4-9}
     & EyePACS & OIA-ODIR & ORIGA & BEH & FIVES & sjchoi86 & REFUGE1 & PAPILA \\ 
    \midrule
    Non-glaucoma & 0 & 4151 & 482 & 463 & 250 & 300 & 720 & 170 \\
    Glaucoma     & 3269 & 291 & 168 & 171 & 150 & 101 & 80 & 47 \\
    \midrule
    Total        & 3269 & 4442 & 650 & 634 & 400 & 401 & 800 & 217 \\
    \bottomrule
    \end{tabular}
    }    
\end{table}

\section{Experiments}
\label{sec:experiment}
\subsection{Datasets}
We validate the effectiveness of the FunOTTA on two eye diseases: DR and glaucoma. For DR, we include six publicly available datasets, namely Kaggle~\cite{kaggle}, Messidor-2~\cite{messidor}, APTOS~\cite{aptos2019}, IDRID~\cite{idrid}, OIA-DDR~\cite{oia-ddr}, and DeepDRiD~\cite{deepdrid}. Each dataset typically includes images classified into categories, including normal, mild, moderate, severe, and proliferative cases. Following the previous domain transfer methods for DR classification, we use Kaggle for training while employing others for testing. 
During training, we randomly downsample all training samples to 708 instances (the smallest class in Kaggle contains 708 samples) per class to alleviate the class imbalance inherent in the Kaggle dataset. However, no downsampling is applied during inference, and all test-time evaluations are conducted on the original, imbalanced datasets to reflect real-world deployment.
For glaucoma, we select a total of eight publicly available datasets from the Standardized Multi-Channel Dataset for Glaucoma (SMDG)~\cite{kiefer2023smdg} based on their data size and quality, namely EyePACS, OIA-ODIR, ORIGA, BEH, FIVES, sjchoi86, REFUGE1, and PAPILA. Each dataset consists of two classes: glaucoma and non-glaucoma. Due to the absence of normal cases in EyePACS, we merge EyePACS and OIA-ODIR to create an approximately balanced training set. 
In the experiments, all fundus images are preprocessed involving background cropping, lighting improvement, padding for missing areas, and resizing to 224 × 224 pixels.

The specific data distribution of both DR and glaucoma is presented in Tab.~\ref{tab:statistics}. 
Fig.~\ref{fig:data_sample} presents sample images and the corresponding RGB statistics from the different sites, where we can observe notable domain-specific styles characterized by variations in color, brightness, and contrast. These domain shifts can be attributed to a combination of factors, including differences in imaging equipment, acquisition protocols, and demographic characteristics. For instance, in the case of DR datasets, the Kaggle dataset was acquired in the USA, APTOS and IDRiD in India, OIA-DDR and DeepDRiD in China, and Messidor-2 in France. Similarly, for glaucoma datasets, EyePACS was collected in the USA, OIA-DDR in China, ORIGA in Singapore, BEH in Bangladesh, REFUGE1 in China, PAPILA in Spain, FIVES in France, and sjchoi86 in Korea.

\input{comp_tab_dr}
\input{comp_tab_g}

\subsection{Evaluation Metrics}
In our study, due to the class imbalance of the target-domain datasets, we select the area under the receiver operating characteristic (ROC) curve, referred to as the AUC, along with the F1 score, to evaluate the performance of TTA methods for DR and glaucoma classification. We conduct experiments to compare the FunOTTA against existing state-of-the-art (SOTA) methods and implement ablation studies to confirm the effectiveness of different components. The AUC and F1 scores for each dataset, as well as the average scores across all datasets, are presented respectively.

\subsection{Implementation Details}
Our experiments are implemented using the PyTorch deep learning framework on one NVIDIA 4090 GPU. We adhere to the rigorous training and evaluation implementation established by \texttt{DomainBed}~\cite{gulrajani2021domainbed} to ensure a fair comparison among TTA methods. During source training, we incorporate data augmentations such as crops of random sizes and aspect ratios, resizing to 224 × 224 pixels, random horizontal flips, and random rotations. 
For each source domain, we fix random seeds to ensure reproducible data partitioning across different compared methods, following an 80\%/20\% split for training and validation. All target domain data remain intact for adaptation. Experiments are conducted across five independent runs with different seeds.

All the backbone networks in our study are pre-trained on ImageNet-1K. The Adam optimizer is utilized with a learning rate of 1e-4 and a batch size of 64, accompanied by a cosine learning rate scheduler. The total training steps for both diseases are 1000 steps. During testing, we freeze the feature extractor and the batch size is 32. We conduct hyperparameter searches for all methods to ensure a fair comparison and report top-1 AUC and F1 scores of these methods on each target domain averaged over five runs. For our method, we test $M\in\{20, 50, 100, N/A\}$, $R\in\{1, 2, 4, 8\}$, where $N/A$ means restoring all samples in the memory bank. The module number $N$ in the ensemble learner is set to 5. Temperature coefficient $\tau$ in Equation~\ref{eq:class_conditionals} is set to 10. $\lambda_{1}$, and $\lambda_{2}$ in Equation~\ref{eq:overall_objective} are set to 0.2 and 1 respectively. 
The upper bound is implemented when training and testing are done in the same target domain. We adopt 5-fold cross-validation, and the reported performance is averaged over five runs. The model is trained for 10 epochs with a learning rate of 1e-4.
The source code, trained models, {training/validation data splits}, as well as the demonstration of FunOTTA on other tasks are available at https://github.com/Casperqian/FunOTTA.

\section{Results}
\subsection{Comparison with SOTA Methods.}
As shown in Tab.~\ref{tab:compare_dr} and Tab.~\ref{tab:compare_glaucoma}, we compare the FunOTTA with state-of-the-art methods across two diseases: 1) TENT~\cite{wang2020tent} minimizes the entropy to obtain the high-confidence prediction on target domains; 2) Pseudo Label (PL)~\cite{lee2013pseudo} finetunes the model by minimizing the cross-entropy between prediction and high-confidence pseudo label; 3) SHOT~\cite{liang2020shot} combines entropy, diversity regularizer, and pseudo-label loss to update the source model; 4) T3A~\cite{iwasawa2021t3a} creates support sets to adjust class prototypes for training-free adaptation; 5) TAST~\cite{jang2022tast} leverages neighbor information and fine-tunes an adaptation module to update class prototypes. 6) UniDG~\cite{zhang2023unidg} introduces a differentiable memory bank and consistency loss to adapt the source model, with loss based on entropy minimization. 7) DeYO~\cite{lee2024deyo} reveals the limitations of using an entropy only and combines a pseudo-label probability difference to obtain a more reliable pseudo-label to update the source model. 
8) TNN~\cite{tnn2024miccai} exploits the relationship between source prototypes and target samples by utilizing neighborhood information, without requiring parameter fine-tuning.
9) EATA~\cite{niu2022eata} adaptively selects reliable and non-redundant test samples to minimize entropy loss while employing a Fisher regularization to prevent catastrophic forgetting.
10) SAR~\cite{niu2023sar} introduces a self-adaptive refinement strategy that leverages consistency regularization and momentum model ensembling to improve prediction reliability during test time.
All comparison methods, including FunOTTA, adhere to the same experimental setup, ensuring a fair and consistent evaluation across all approaches.
In our study, TENT, PL, SHOT, and DeYO are slightly modified due to the high sensitivity of the medical diagnosis model to parameter changes. Specifically, TENT, PL, and DeYO update the batch normalization layers, while SHOT modifies the feature extractor, as described in the original literature. We observe poor performance when reproducing these methods in our setting. However, by only optimizing the classifier weights, we achieve significant improvement.

Tab.~\ref{tab:compare_dr} and Tab.~\ref{tab:compare_glaucoma} present the comparison results. 
To determine statistical significance, we conduct paired t-tests between FunOTTA and the compared methods. To reduce the number of comparisons, we perform the t-tests with the second performing methods. For each metric (AUC or F1), the tests are performed on 25 data pairs from the DR datasets (5 domains $\times$ 5 runs) and 30 from the glaucoma datasets (6 domains $\times$ 5 runs), with a p-value $<$ 0.05 denoting statistical significance.
The upper bound refers to the performance achieved when both training and testing are conducted on the target domain. The performance gap between the upper bound and the baseline highlights the significant degradation caused by domain shift across all evaluation metrics.
Our FunOTTA framework outperforms others across two diseases, a total of 11 target domains. Besides ResNet, we also evaluate EfficientNet and DenseNet as our backbone networks. However, since ResNet is widely adopted in fundus image classification, we choose to focus on its results. For the five diabetic retinopathy datasets, our method shows an average improvement of 2.0 in AUC and 6.5 in F1 score compared to the backbone ResNet18, and an improvement of 3.1 in AUC and 10.4 in F1 score compared to the backbone ResNet50. For the six glaucoma datasets, our method shows an average improvement of 3.2 in AUC and 8.3 in F1 score compared to the backbone ResNet18, and an improvement of 1.4 in AUC and 4.1 in F1 score compared to the backbone ResNet50. Remarkably, our FunOTTA framework, with a training-based setting, is the first TTA method for fundus images that not only maximizes the performance of training-based approaches but also demonstrates exceptional robustness.

Notably, except for certain domains like Messidor-2 when using ResNet18, where most TTA methods exhibit a decrease in performance compared to the ERM baseline, the FunOTTA consistently outperforms all state-of-the-art methods across multiple target domains. Messidor-2 may have a relatively stronger data bias compared to the other datasets, which could explain this result. Since the performance variation across five runs for each dataset remains within a small range (all below 0.7), we do not display them in the results. Many training-based TTA methods yield lower performance than ERM baselines, consistent with the findings in~\cite{tnn2024miccai}.
Among them, UniDG degrades the most on both diseases, likely due to its full-parameter updates during adaptation, which can harm the parameter-sensitive models despite the use of consistency loss. Its alignment of intermediate features also makes partial updates difficult.
Methods relying heavily on prediction entropy minimization, such as TENT and SHOT, also exhibit performance drops. This is because entropy is often noisy in the context of fundus images, leading to suboptimal adaptation decisions. In contrast, EATA and SAR achieve better performance since they leverage selective entropy during adaptation. Instead of fully trusting entropy, they incorporate strategies to identify and exclude uncertain predictions.
Notably, DeYO achieves substantial improvements in certain domains; however, its object-destructive augmentations may distort critical pathological patterns, limiting its applicability in fundus image analysis.
Similarly, TNN provides stable performance and highlights the strength of dynamic neighbor voting for robust adaptation. Nonetheless, its reliance on frozen parameters constrains the upper bound of adaptation.

Additionally, in the Glaucoma datasets, most TTA methods show minor differences in performance compared to the DR datasets. It should be attributed to the inter-class semantic gaps. Specifically, DR grading relies on identifying lesions such as retinal detachment, exudate, and hemorrhages~\cite{feldman2019diabetic}, whereas glaucoma diagnosis depends on subtle features like the optic cup-to-disc ratio, which is smaller and more complex to distinguish~\cite{stein2021glaucoma}. This observation aligns with our statement that smaller inter-class semantic gaps pose great challenges for TTA.

\begin{figure*}[htbp]
    \centering
    \includegraphics[width=\linewidth]{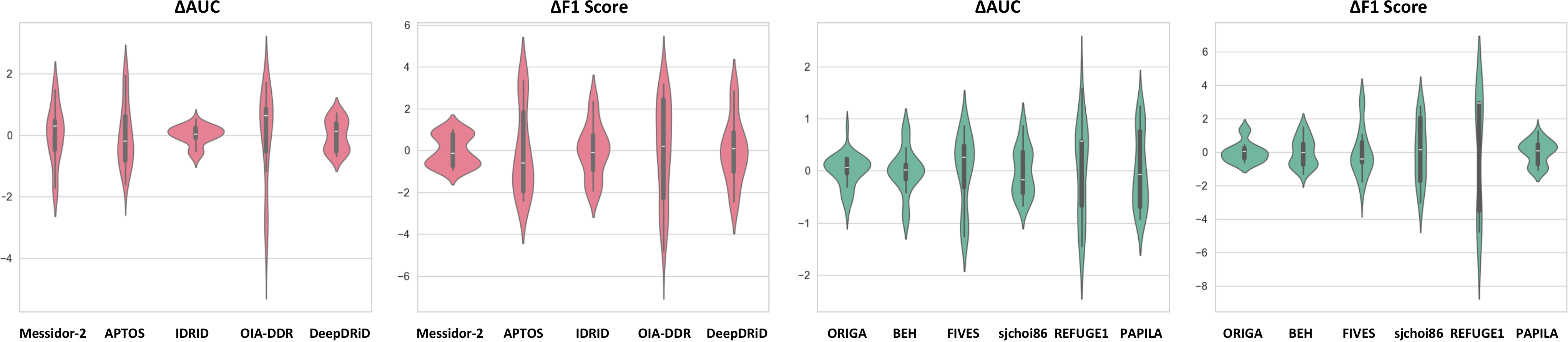}
    \caption{Distribution of performance improvements by the FunOTTA for models trained with different hyperparameters.}
    \label{fig:distribution_variation}
\end{figure*}

\begin{figure}[htbp]
    \centering
    \subfloat{\includegraphics[width=0.5\linewidth]{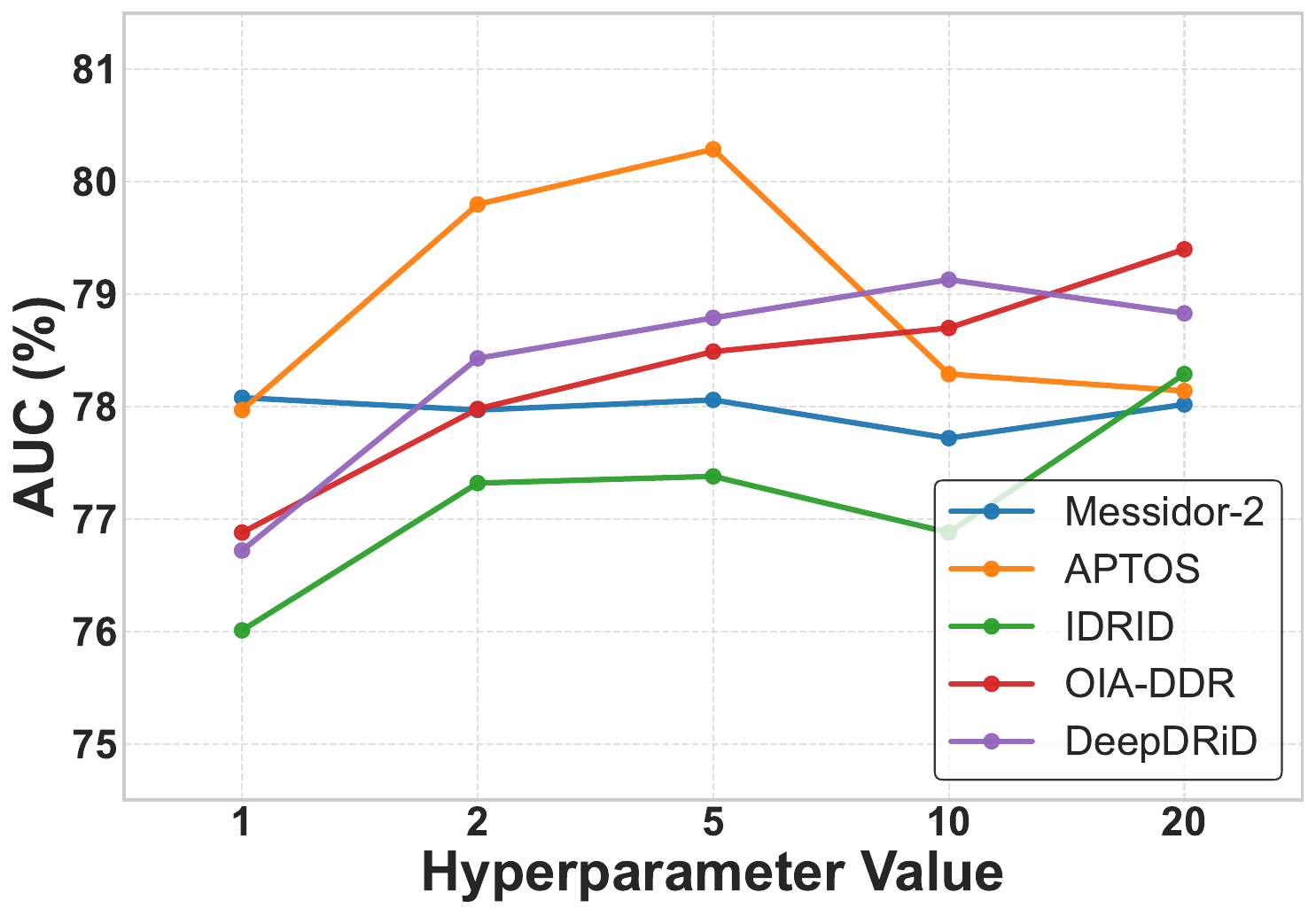}}
    \subfloat{\includegraphics[width=0.5\linewidth]{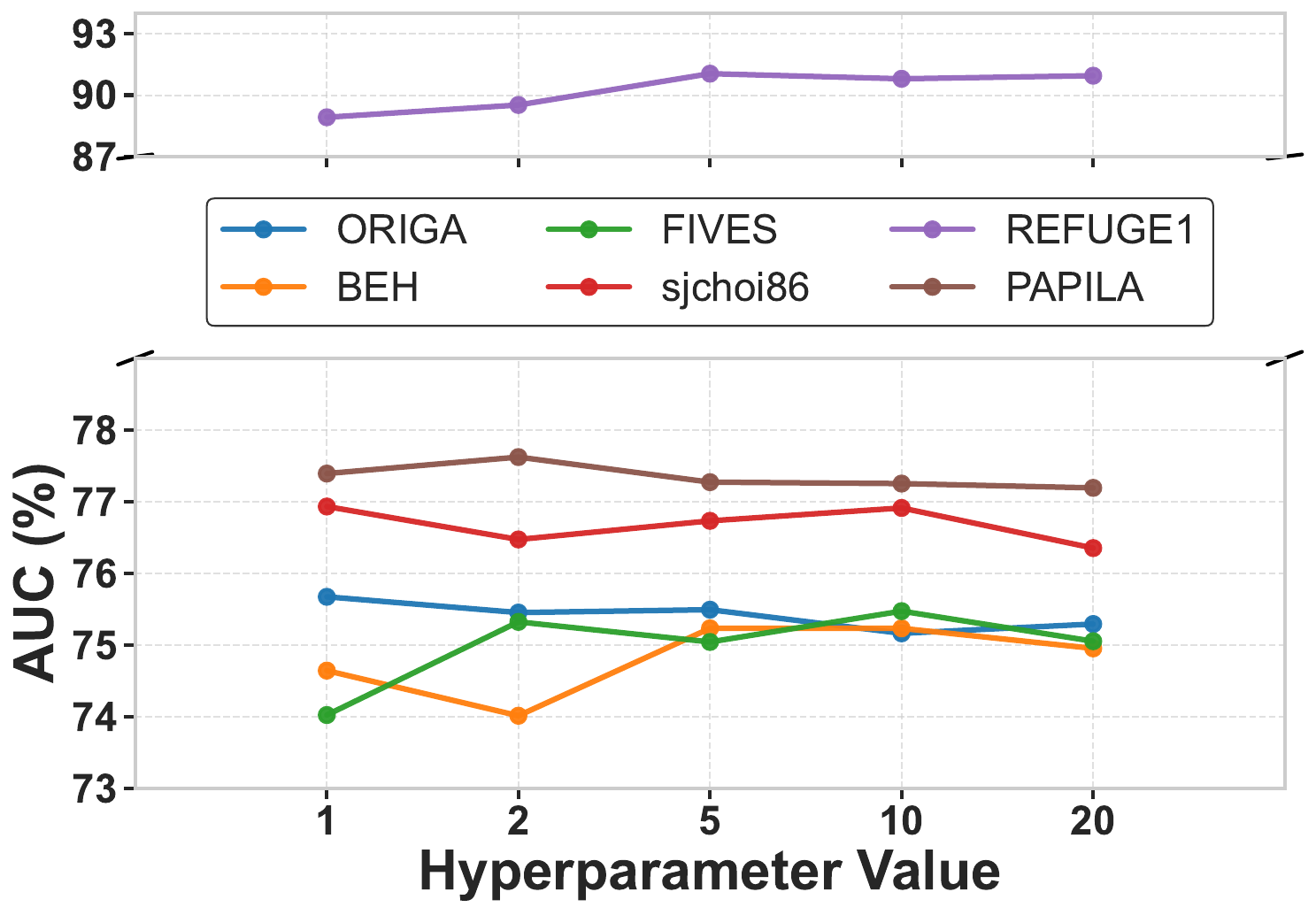}}
    \caption{Effect of the number of ensemble learners across DR and glaucoma datasets. Here we used ResNet50 as the backbone network.}
    \label{fig:hyper_ensemble}
\end{figure}

\begin{figure*}[htbp]
    \centering
    \includegraphics[width=\linewidth]{pic/ablation_mem.pdf}
    \caption{Comparison between different filtering mechanisms across two diseases. The results are averaged across five runs, using ResNet50 as the backbone network.}
    \label{fig:entropy_vs_ours}
\end{figure*}

\subsection{Stability in Hyperparameter Selection}
We also explore the stability of performance improvements when selecting different hyperparameters. Fig.~\ref{fig:distribution_variation} shows the distribution of performance improvements of the FunOTTA across various hyperparameter settings. Specifically, we evaluate performance under a range of values for $M\in\left\{20, 50, 100\right\}$ and $R\in\left\{1, 2, 4, 8\right\}$, with only one optimization step applied to each incoming batch, and trial for five runs. The results indicate that, although our method is affected by hyperparameter choices to a certain extent, particularly in domains like OIA-DDR and REFUGE1, it demonstrates strong robustness, as the variation between the best and worst outcomes remains relatively small. In contrast, T3A, even on computer vision domains with clearer semantic gaps (e.g., VLCS~\cite{fang2013unbiased}, PACS~\cite{li2017deeper}), exhibits a broader fluctuation of approximately -10\% to 10\% when adjusting its hyperparameters~\cite{iwasawa2021t3a}.

We further analyze the influence of the number of ensemble learners $N$ on model performance. In addition to the default setting, we evaluate a wide range of values with $N \in \left\{1, 2, 5, 10, 20\right\}$ across multiple target domains. As illustrated in Fig.~\ref{fig:hyper_ensemble}, increasing the number of learners from a single model ($N=1$) to multiple learners leads to a noticeable performance improvement. However, the gain becomes marginal beyond $N=5$: the curves for $N=5$, $10$, and $20$ are nearly overlapping, indicating a saturation effect. These results suggest that ensembles consistently outperform single models, but increasing $N$ beyond a moderate size yields diminishing returns. To strike a balance between accuracy and computational efficiency, we recommend setting $N=5$ as the default. This choice provides robust performance without requiring further task-specific tuning.

\begin{figure*}[th]
    \centering
    \subfloat{
    \includegraphics[width=0.5\linewidth]{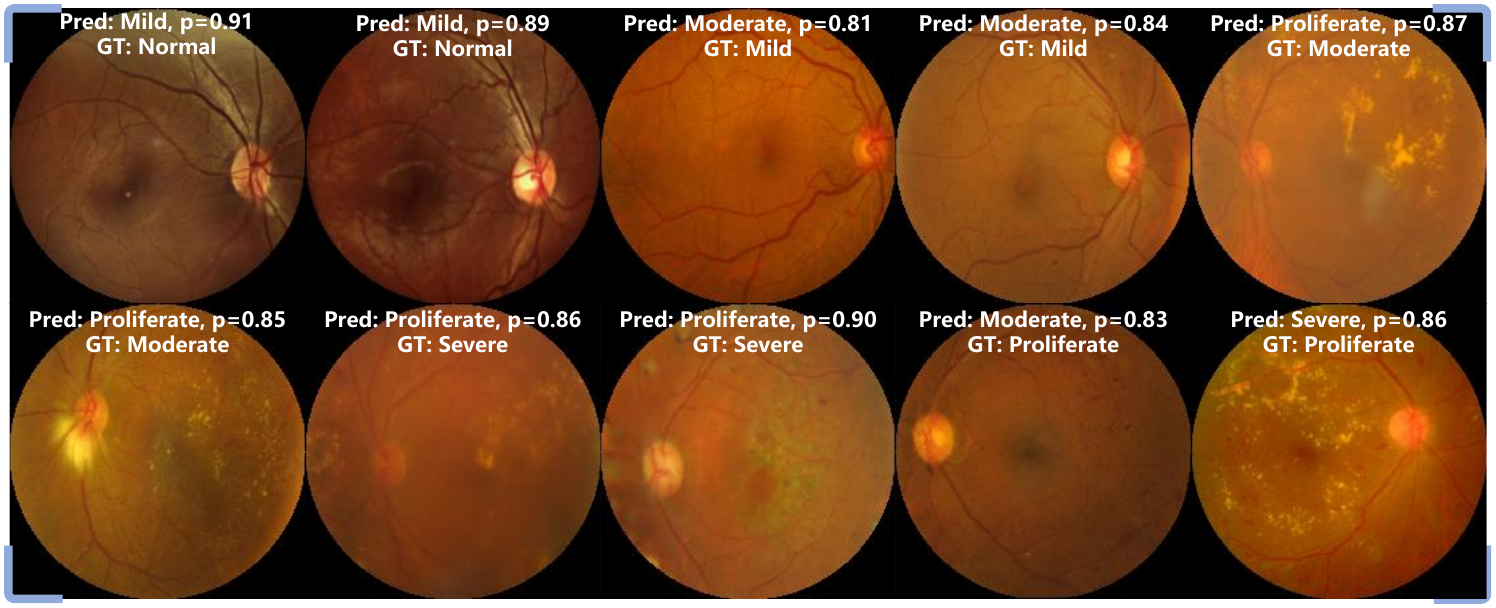}}
    \subfloat{
    \includegraphics[width=0.5\linewidth]{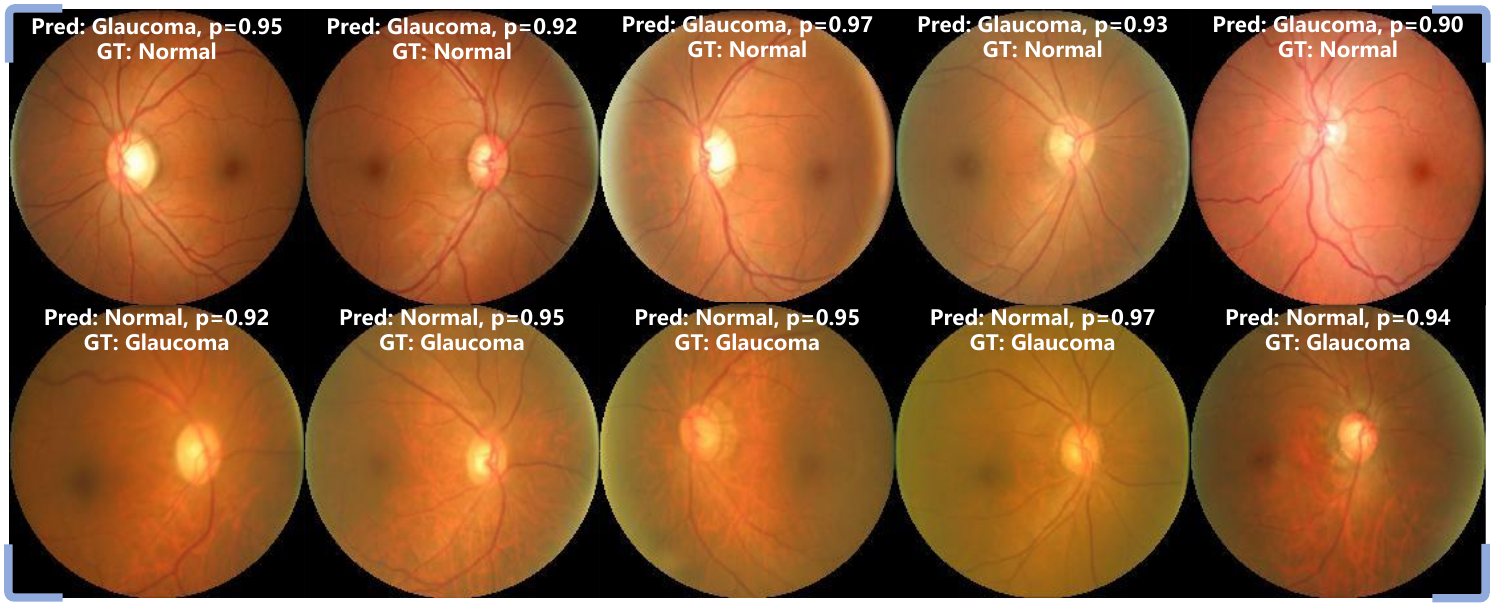}}
    \caption{Fundus images from DR (left) and glaucoma (right) datasets misclassified by ResNet50 but correctly classified by FunOTTA. ResNet50 predictions with confidence score, as well as the ground truth are shown in white.}
    \label{fig:visualization}
\end{figure*}

\input{ablation_tab}

\subsection{Ablation Study}
We conduct ablation studies to evaluate the contributions of our proposed components in enhancing performance. 
As shown in Fig.~\ref{fig:entropy_vs_ours}, we compare the results of our dynamic filtering mechanism with conventional entropy-based filtering. 
Our approach consistently outperforms the conventional method across all datasets. These results confirm that, rather than relying on noisy entropy, our dynamic filtering mechanism facilitates the extraction of more reliable instance representations, thereby effectively suppressing harmful knowledge bias introduced by the source predictor.
Notably, in terms of the F1 score, our dynamic filtering mechanism achieves significant improvements on multiple datasets, validating the importance of filtering out ambiguous latent features for better adaptation.

As shown in Tab.~\ref{tab:ablation_dr} and Tab.~\ref{tab:ablation_glaucoma}, the combined use of $\mathcal{L}_{CCL}$ and $\mathcal{L}_{DAL}$ results in performance gains in target domains compared to using either one loss alone or neither. Specifically, compared to the baseline without any loss, our method achieves a 3.35 increase in AUC and a 6.62 increase in F1 score on DR datasets, as well as a 1.99 increase in AUC and a 2.61 increase in F1 score on glaucoma datasets. These results highlight the effectiveness of our proposed components in addressing domain shifts inherent in fundus images.

\begin{figure}[h]
    \centering
    \includegraphics[width=\linewidth]{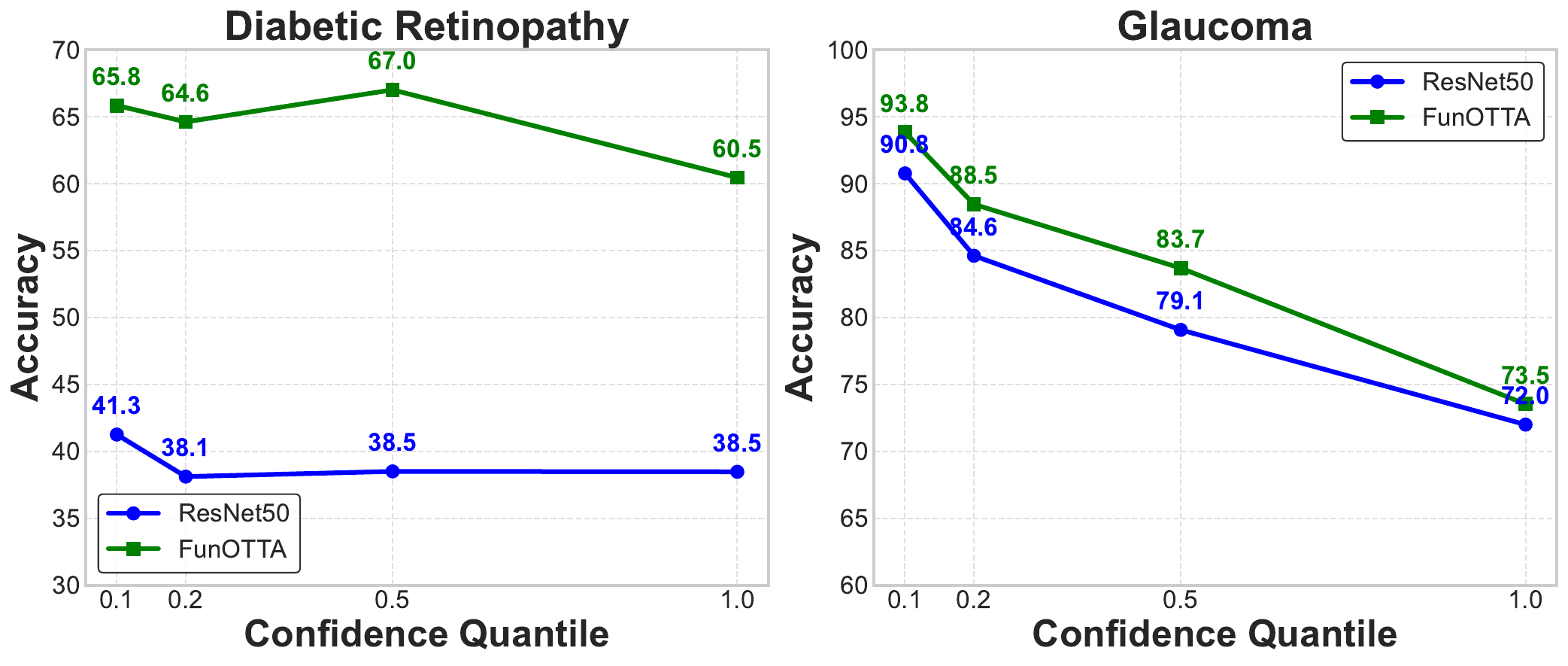}
    \caption{The relation between accuracy and confidence quantile for ATPOS (left) and ORIGA (right).}
    \label{fig:quantile}
\end{figure}

\subsection{Managing Entropy for Reliable TTA}
\label{sec:Managing Entropy for Reliable TTA}
We further investigate why entropy can sometimes negatively impact TTA in the context of fundus images. Specifically, we focus on the APTOS and ORIGA datasets, as FunOTTA demonstrates superior performance in these domains for DR grading and glaucoma detection. In Fig.~\ref{fig:visualization}, we present examples of fundus images misclassified by ResNet50 with high confidence, taken from the APTOS and ORIGA. Remarkably, FunOTTA, by reducing reliance on potentially harmful prior knowledge, correctly classifies these images into the appropriate categories. These results suggest that part of the entropy value, representing prediction confidence, can mislead the model during TTA by conveying unreliable yet high-confidence information. We term such samples with high-confidence erroneous predictions as challenging samples. Conventional TTA tends to memorize the information from these samples, ultimately leading to limited performance.

Fig.~\ref{fig:quantile} explores the role of entropy in extracting useful content in target domains. The confidence quantile represents the subset of fundus images with the lowest entropy. For example, a confidence quantile of 0.1 means only the subset with the lowest 10\% of entropy is considered. Notably, we use accuracy as the evaluation metric rather than AUC or F1 scores, as it directly reflects the impact of unreliable entropy in traditional TTA operations, such as entropy minimization or prototype construction.
In the APTOS domain, accuracy does not consistently improve as the confidence quantile decreases, indicating that entropy is not reliably informative in this domain. In contrast, in the ORIGA domain, accuracy shows a clear inverse relationship with the confidence quantile, suggesting that lower entropy correlates with trustworthy samples in this context. This highlights why traditional TTA methods often fail in the APTOS domain, as they heavily rely on entropy, which is less reliable there. Conversely, the same TTA methods perform well in ORIGA, where entropy conveys valuable information. Notably, FunOTTA achieves impressive performance on both APTOS and ORIGA because it effectively manages entropy information. By reducing reliance on harmful entropy, FunOTTA prevents collapse in entropy-sensitive domains. At the same time, it minimizes the use of confident but unreliable predictions, which improves the upper bound in domains like ORIGA.

Based on these findings, we can conclude that addressing incorrect but confident predictions is crucial for both the lower and upper bounds of TTA methods. Effectively managing and utilizing entropy, especially in entropy-sensitive domains, is key to optimizing TTA methods.

\subsection{Further Analysis on Label Shift}
\begin{figure}[t]
    \centering
    \includegraphics[width=.9\linewidth]{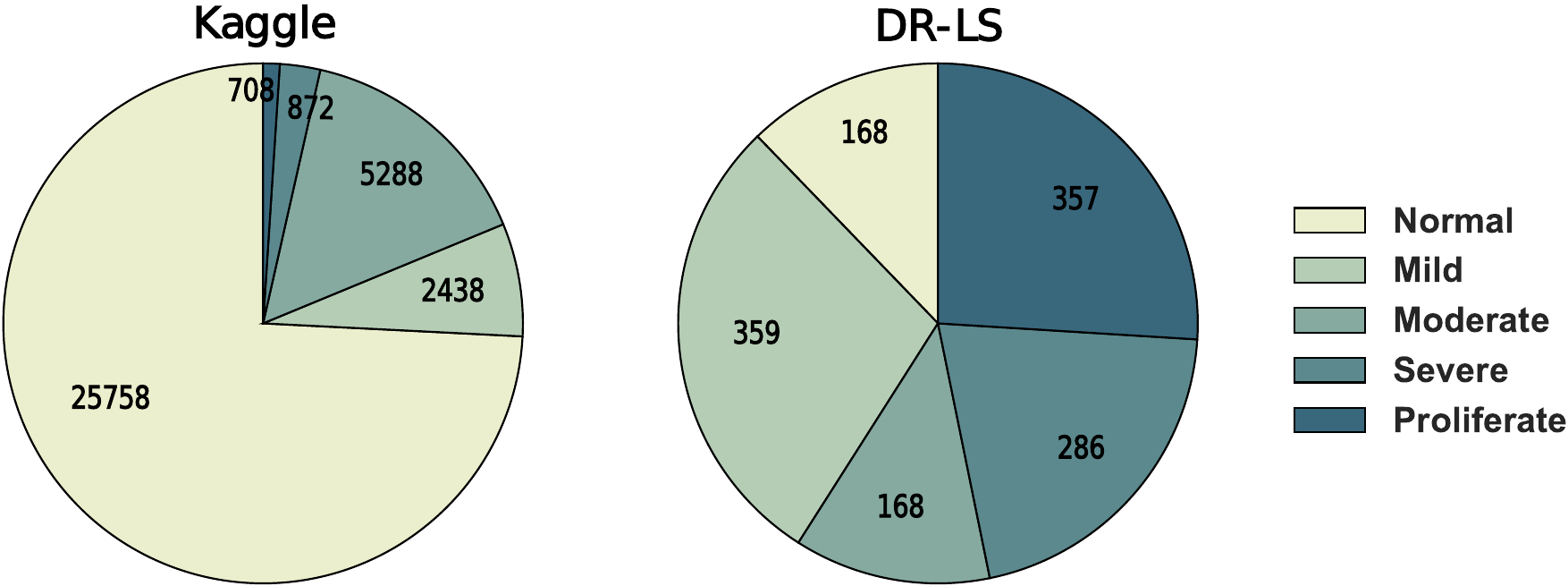}
    \caption{{Data distribution of the Kaggle and label-shifted DR (DR-LS) datasets.}}
    \label{fig:DR-LS}
\end{figure}
\input{comp_tab_ls}

We further analyze FunOTTA under both covariate shifts and label shifts. Since existing datasets share similar label-space distributions, we construct a curated label-shifted DR dataset (DR-LS) by merging the mild, severe, and proliferate samples from APTOS into IDRiD, while preserving the remaining categories from IDRiD, as these two datasets exhibit similar RGB statistics (see Fig.~\ref{fig:data_sample}). To evaluate model performance, we train the source model on the original Kaggle dataset and test it on DR-LS. The detailed data distribution is presented in Fig.~\ref{fig:DR-LS}.
Tab.~\ref{tab:compare_ls} reports the comparison results under this label-shift scenario. FunOTTA consistently surpasses existing baselines and demonstrates strong robustness under both covariate and label shifts, highlighting its capability to transfer effectively to scenarios with shifted label distributions.

Beyond conventional label-shift settings, more challenging scenarios arise when the target domain contains classes unseen during source training. Under this setting, universal domain adaptation (UniDA) aims to handle situations where the label spaces of source and target domains are disjoint~\cite{you2019unida}. Recent UniDA methods, such as DANCE~\cite{saito2020dance} and MemSPM
~\cite{lai2023MemSPM}, also incorporate popular strategies like entropy regularization and memory bank to align distributions. Notably, the design of FunOTTA is conceptually aligned with these principles and further advances them, enabling natural transfer to more challenging settings. At the same time, its unique and more effective design such as dynamic filtering mechanism allows FunOTTA to better exploit these principles, potentially achieving stronger adaptation to unseen classes. This indicates that FunOTTA holds high potential to handle more extreme domain shifts involving unseen classes, further underscoring its adaptability.

\section{Conclusion}
\label{sec:conclution}
In this paper, we introduce a novel FunOTTA framework based on stable feature learning, designed to adapt off-the-shelf fundus image diagnosis models to unseen domains. We verify the effectiveness of the FunOTTA using extensive experimental results and demonstrate that the proposed method achieves state-of-the-art performance on large-scale fundus image benchmarks. To the best of our knowledge, this is the first application of training-based TTA for fundus image diagnosis, successfully addressing the common issue of performance degradation often encountered by training-based methods in fundus imaging. 
Additionally, our method offers the potential for seamless extension to other medical scenarios with distribution shifts, establishing a foundation for future applications of TTA in medical image analysis (see additional applications in our code repository).

The limitations of the FunOTTA and potential directions for future work are as follows.
First, our method maintains the close-world assumption in the label space. We will further address this limitation by extending evaluations to datasets involving rare diseases, thereby enhancing generalization under open-set scenarios.
Second, hyperparameter tuning remains a practical challenge in real-world medical applications, where exhaustive search is often infeasible. In future work, we aim to explore more adaptive strategies, such as leveraging attention mechanisms in neighbor selection, to automatically determine the appropriate number of neighbors without manual tuning.
We hope that the findings of this paper will contribute to developing more effective test-time adaptation methods in medical contexts.




\input{ref.bbl}

\end{document}

%% file: comp_tab_dr.tex
\begin{table*}[th]
\caption{Comparison of methods across domains in diabetic retinopathy. Our results are averaged over five runs. The dagger symbol ($\dag$) denotes that we implemented the method by only fine-tuning the classifier weights, as the original approach performed poorly on fundus images. The highest values are highlighted in \textbf{bold}, the second values are \underline{underlined}, {and * denotes statistical significance in paired t-test between FunOTTA and the second performing method (** indicates $p\leq0.01$, * indicates $p\leq0.05$).}}
\label{tab:compare_dr}
\centering
\resizebox{0.85\linewidth}{!}{
\setlength{\tabcolsep}{3pt}
\begin{tabular}{l|cc|cc|cc|cc|cc|cc}
\toprule
{Dataset} & \multicolumn{2}{c|}{{Messidor-2}} & \multicolumn{2}{c|}{{APTOS}} & \multicolumn{2}{c|}{{IDRID}} & \multicolumn{2}{c|}{{OIA-DDR}} & \multicolumn{2}{c|}{{DeepDRiD}} & \multicolumn{2}{c}{{Average}} \\ 
\midrule
{Metric}  & {AUC} & {F1} & {AUC} & {F1} & {AUC} & {F1} & {AUC} & {F1} & {AUC} & {F1} & {AUC} & {F1} \\ 
\midrule
ResNet18 
&75.9{$_{0.4}$} &51.0{$_{0.7}$} 
&73.5{$_{1.8}$} &45.0{$_{2.8}$} 
&77.0{$_{0.9}$} &36.6{$_{3.5}$} 
&76.2{$_{0.5}$} &44.5{$_{1.2}$} 
&76.7{$_{1.4}$} &45.6{$_{2.5}$} 
&75.9 &44.6 \\

+TENT$^{\dag}$~\cite{wang2020tent}
&75.8{$_{0.6}$} &\textbf{52.6}{$_{0.9}$} 
&\underline{73.8}{$_{1.6}$} &51.3{$_{9.6}$} 
&76.7{$_{0.9}$} &34.3{$_{3.7}$} 
&74.8{$_{0.9}$} &51.4{$_{7.9}$} 
&76.9{$_{1.7}$} &47.4{$_{2.0}$} 
&75.6 &47.4 \\

+PL$^{\dag}$~\cite{lee2013pseudo}
&75.8{$_{0.2}$} &51.5{$_{0.8}$} 
&71.2{$_{1.8}$} &48.1{$_{3.5}$} 
&77.0{$_{0.9}$} &36.3{$_{3.6}$} 
&73.9{$_{1.1}$} &46.0{$_{3.7}$} 
&76.5{$_{1.4}$} &45.2{$_{2.4}$} 
&74.9 &45.4 \\

+SHOT$^{\dag}$~\cite{liang2020shot}
&75.8{$_{0.5}$} &50.6{$_{0.6}$} 
&73.3{$_{2.3}$} &47.0{$_{6.8}$} 
&78.4{$_{0.9}$} &\underline{43.1}{$_{1.8}$} 
&76.0{$_{0.7}$} &43.3{$_{0.9}$} 
&78.2{$_{0.7}$} &49.2{$_{0.7}$} 
&76.3 &46.6 \\

+T3A~\cite{iwasawa2021t3a} 
&73.5{$_{0.9}$} &42.7{$_{1.2}$} 
&71.6{$_{3.9}$} &48.0{$_{6.2}$} 
&78.2{$_{1.1}$} &37.3{$_{1.9}$} 
&77.6{$_{0.9}$} &48.7{$_{3.1}$} 
&\underline{79.3}{$_{1.8}$} &47.8{$_{2.4}$} 
&76.0 &44.9 \\
 
+TAST~\cite{jang2022tast}
&67.9{$_{1.8}$} &45.7{$_{2.3}$} 
&72.8{$_{4.2}$} &49.8{$_{7.1}$} 
&75.8{$_{1.4}$} &42.2{$_{1.5}$} 
&71.6{$_{0.7}$} &47.1{$_{2.1}$} 
&77.4{$_{1.8}$} &\underline{50.2}{$_{1.5}$} 
&73.1 &47.0 \\

+UniDG~\cite{zhang2023unidg}
&67.2{$_{3.3}$} &47.2{$_{2.0}$} 
&73.2{$_{9.3}$} &\underline{53.4}{$_{4.3}$} 
&68.7{$_{6.8}$} &32.9{$_{5.4}$} 
&76.0{$_{2.2}$} &\textbf{58.2}{$_{5.0}$} 
&69.3{$_{1.1}$} &47.3{$_{0.9}$} 
&70.9 &47.8 \\

+DeYO$^{\dag}$~\cite{lee2024deyo}
&74.8{$_{2.2}$} &48.1{$_{2.2}$} 
&71.2{$_{1.5}$} &45.1{$_{1.0}$} 
&\underline{78.5}{$_{1.0}$} &39.0{$_{2.5}$} 
&70.0{$_{0.9}$} &41.7{$_{2.7}$} 
&75.8{$_{2.1}$} &48.5{$_{1.6}$} 
&74.0 &43.6 \\

+TNN~\cite{tnn2024miccai} & 75.0{$_{1.0}$} & 50.4{$_{1.3}$} & 72.5{$_{3.2}$} & 49.2{$_{5.9}$} & 77.2{$_{1.3}$} & 41.4{$_{1.7}$} & \underline{78.1}{$_{1.2}$} & 49.2{$_{3.2}$} & 79.0{$_{0.9}$} & 48.6{$_{1.4}$} & 76.1 & \underline{48.0} \\

+EATA$^{\dag}$~\cite{niu2022eata}
&\textbf{76.0}{$_{0.5}$} &51.1{$_{1.0}$} 
&73.5{$_{1.9}$} &49.0{$_{2.6}$} 
&77.1{$_{0.9}$} &40.6{$_{4.0}$} 
&77.2{$_{0.6}$} &46.4{$_{3.6}$} 
&77.1{$_{1.4}$} &47.6{$_{2.5}$} 
&76.2 &46.9 \\

+SAR$^{\dag}$~\cite{niu2023sar}
&\underline{75.9}{$_{0.5}$} &51.8{$_{0.8}$} 
&74.9{$_{1.0}$} &50.1{$_{4.0}$} 
&77.9{$_{1.0}$} &41.4{$_{3.9}$} 
&76.1{$_{0.7}$} &46.7{$_{2.3}$} 
&78.0{$_{1.4}$} &48.5{$_{2.3}$} 
&\underline{76.6} &47.7 \\

+FunOTTA
&74.0{$_{2.6}$} &\underline{52.0}{$_{3.3}$} 
&\textbf{77.1}{$_{3.3}$} &\textbf{55.4}{$_{5.5}$} 
&\textbf{79.0}{$_{1.2}$} &\textbf{43.9}{$_{1.7}$} 
&\textbf{79.5}{$_{0.5}$} &\underline{51.6}{$_{1.6}$} 
&\textbf{79.8}{$_{1.3}$} &\textbf{52.2}{$_{1.8}$} 
&\textbf{77.9}{$^{*}$} &\textbf{51.0}{$^{**}$} \\
\midrule
Upper bound & 81.3{$_{1.1}$} & 58.2{$_{1.3}$} & 87.9{$_{1.6}$} & 70.8{$_{2.0}$} & 82.9{$_{1.5}$} & 61.9{$_{3.3}$} & 86.6{$_{1.2}$} & 71.4{$_{1.5}$} & 86.8{$_{1.7}$} & 64.8{$_{2.8}$} & 84.72 & 65.20 \\
\midrule
ResNet50
&78.9{$_{2.3}$} &55.6{$_{4.6}$} 
&73.9{$_{2.0}$} &42.5{$_{7.0}$} 
&74.9{$_{1.2}$} &27.2{$_{1.9}$} 
&77.0{$_{0.8}$} &42.2{$_{0.6}$} 
&74.9{$_{0.8}$} &37.0{$_{0.6}$} 
&75.9 &40.9 \\

+TENT$^{\dag}$~\cite{wang2020tent}
&\underline{79.0}{$_{2.3}$} &55.1{$_{0.8}$} 
&69.8{$_{1.0}$} &47.3{$_{8.9}$} 
&74.6{$_{1.1}$} &26.5{$_{2.1}$} 
&73.4{$_{0.4}$} &25.4{$_{5.8}$} 
&74.7{$_{0.9}$} &34.9{$_{1.1}$} 
&74.3 &37.8 \\

+PL$^{\dag}$~\cite{lee2013pseudo}
&78.9{$_{2.2}$} &55.1{$_{4.9}$} 
&71.7{$_{0.9}$} &41.4{$_{8.6}$} 
&75.6{$_{1.1}$} &26.9{$_{1.3}$} 
&75.3{$_{2.7}$} &45.2{$_{9.0}$} 
&74.3{$_{0.9}$} &36.1{$_{2.2}$} 
&75.2 &40.9 \\

+SHOT$^{\dag}$~\cite{liang2020shot}
&78.8{$_{2.3}$} &54.3{$_{3.1}$} 
&73.9{$_{3.3}$} &44.1{$_{4.9}$} 
&76.6{$_{0.7}$} &\textbf{42.1}{$_{2.1}$} 
&75.7{$_{0.6}$} &42.5{$_{0.8}$} 
&76.4{$_{0.5}$} &40.8{$_{1.1}$} 
&76.3 &44.4 \\

+T3A~\cite{iwasawa2021t3a}
&77.8{$_{2.2}$} &52.4{$_{4.0}$} 
&\underline{75.9}{$_{5.7}$} &44.2{$_{5.2}$} 
&76.2{$_{1.3}$} &33.6{$_{0.8}$} 
&\underline{78.7}{$_{1.8}$} &47.7{$_{1.9}$} 
&76.8{$_{0.2}$} &40.0{$_{0.3}$} 
&77.1 &43.6 \\

+TAST~\cite{jang2022tast}
&77.3{$_{4.0}$} &55.4{$_{3.5}$} 
&74.0{$_{4.9}$} &\underline{54.1}{$_{6.6}$} 
&71.7{$_{3.6}$} &35.9{$_{1.9}$} 
&71.9{$_{1.3}$} &47.3{$_{0.8}$} 
&75.2{$_{1.4}$} &43.2{$_{2.2}$} 
&74.0 &47.2 \\

+UniDG~\cite{zhang2023unidg}
&68.6{$_{1.6}$} &43.6{$_{0.2}$} 
&70.6{$_{4.0}$} &46.8{$_{9.6}$} 
&68.3{$_{5.0}$} &30.9{$_{9.5}$} 
&73.9{$_{2.5}$} &\underline{49.7}{$_{6.3}$} 
&62.6{$_{1.3}$} &31.8{$_{4.1}$} 
&68.8 &40.6 \\

+DeYO$^{\dag}$~\cite{lee2024deyo}
&73.8{$_{2.1}$} &50.1{$_{0.2}$} 
&68.9{$_{3.5}$} &48.1{$_{8.4}$} 
&76.8{$_{1.6}$} &32.9{$_{4.0}$} 
&73.9{$_{3.7}$} &33.8{$_{2.9}$} 
&75.9{$_{0.7}$} &\underline{47.5}{$_{3.5}$} 
&73.9 &42.5 \\

+TNN~\cite{tnn2024miccai} & 79.0{$_{2.4}$}  & 55.3{$_{3.9}$}  & 75.2{$_{4.7}$}  & 53.9{$_{5.0}$}  & \underline{77.0}{$_{1.0}$}  & 36.4{$_{1.8}$}  & 77.6{$_{2.0}$}  & 48.4{$_{2.3}$}  & \underline{77.0{$_{0.8}$} } & 43.1{$_{1.7}$}  & \underline{77.2} & \underline{47.4} \\

+EATA$^{\dag}$~\cite{niu2022eata}
&78.8{$_{2.5}$} &\underline{55.9}{$_{2.1}$} 
&73.9{$_{1.1}$} &47.5{$_{1.8}$} 
&75.1{$_{1.1}$} &35.1{$_{1.8}$} 
&77.5{$_{0.7}$} &44.4{$_{0.7}$} 
&75.8{$_{0.9}$} &38.5{$_{0.7}$} 
&76.2 &44.3 \\

+SAR$^{\dag}$~\cite{niu2023sar}
&78.4{$_{2.4}$} &54.6{$_{4.5}$} 
&74.6{$_{1.2}$} &48.2{$_{2.3}$} 
&76.8{$_{1.5}$} &35.3{$_{2.8}$} 
&78.2{$_{0.9}$} &45.0{$_{1.2}$} 
&75.6{$_{2.1}$} &39.8{$_{2.7}$} 
&76.7 &44.6 \\

+FunOTTA
&\textbf{79.1}{$_{2.8}$} &\textbf{56.3}{$_{1.6}$} 
&\textbf{81.9}{$_{3.0}$} &\textbf{60.0}{$_{2.8}$} 
&\textbf{77.6}{$_{1.9}$} &\underline{40.1}{$_{2.2}$} 
&\textbf{79.0}{$_{1.0}$} &\textbf{51.8}{$_{6.0}$} 
&\textbf{77.2}{$_{0.9}$} &\textbf{48.4}{$_{3.3}$} 
&\textbf{79.0}{$^{*}$} & \textbf{51.3}{$^{**}$} \\
\midrule
Upper bound & 83.3{$_{2.4}$} & 66.7{$_{3.4}$} & 89.7{$_{1.4}$} & 74.5{$_{3.7}$} & 81.8{$_{1.3}$} & 60.9{$_{1.6}$} & 87.6{$_{1.0}$} & 74.8{$_{1.2}$} & 85.5{$_{1.9}$} & 61.2{$_{2.1}$} & 85.2 & 68.0 \\
\bottomrule
\end{tabular}
}
\end{table*}

%% file: comp_tab_g.tex
\begin{table*}[ht]
\caption{Comparison of methods across domains in glaucoma. Our results are averaged over five runs. The dagger symbol ($\dag$) denotes that we implemented the method by only fine-tuning the classifier weights, as the original approach performed poorly on fundus images. The highest values are highlighted in \textbf{bold}, the second values are \underline{underlined}, {and * denotes statistical significance in paired t-test between FunOTTA and the second performing method (** indicates $p\leq0.01$, * indicates $p\leq0.05$).}}
\label{tab:compare_glaucoma}
\centering
\resizebox{.95\linewidth}{!}{
\setlength{\tabcolsep}{3pt}
\begin{tabular}{l|cc|cc|cc|cc|cc|cc|ll}
\toprule
{Method} & \multicolumn{2}{c|}{{ORIGA}} & \multicolumn{2}{c|}{{BEH}} & \multicolumn{2}{c|}{{FIVES}} 
& \multicolumn{2}{c|}{{sjchoi86}} & \multicolumn{2}{c|}{{REFUGE1}} & \multicolumn{2}{c|}{{PAPILA}} & \multicolumn{2}{c}{{Average}} \\ 
\midrule
{Metric} & {AUC} & {F1} & {AUC} & {F1} & {AUC} & {F1} & {AUC} & {F1} & {AUC} & {F1} & {AUC} & {F1} & {AUC} & {F1} \\ 
\midrule
{ResNet18}&75.0{$_{1.0}$} &53.7{$_{1.2}$} &72.7{$_{0.7}$} &54.1{$_{1.4}$} &73.5{$_{1.2}$} &45.8{$_{2.1}$} &75.4{$_{2.2}$} &51.9{$_{3.4}$} &87.7{$_{1.1}$} &52.6{$_{3.3}$} &73.8{$_{2.2}$} &43.6{$_{3.1}$} &76.3 &50.3 \\

{+TENT$^{\dag}$~\cite{wang2020tent}} &75.1{$_{1.0}$} &54.7{$_{1.2}$} &72.9{$_{0.4}$} &53.8{$_{1.2}$} &73.5{$_{1.2}$} &42.7{$_{1.9}$} &75.3{$_{2.2}$} &50.9{$_{4.1}$} &87.9{$_{1.1}$} &53.9{$_{3.5}$} &73.4{$_{3.1}$} &43.7{$_{1.3}$} &76.3 &50.0 \\  

{+PL$^{\dag}$~\cite{lee2013pseudo}} 
&75.2{$_{1.0}$} &54.4{$_{1.3}$} 
&73.2{$_{0.4}$} &53.7{$_{1.9}$} 
&73.5{$_{1.2}$} &42.7{$_{2.0}$} 
&75.3{$_{2.2}$} &50.9{$_{3.9}$} 
&87.8{$_{1.1}$} &52.8{$_{3.4}$} 
&73.4{$_{2.8}$} &43.3{$_{1.6}$} 
&76.4 &49.6 \\

{+SHOT$^{\dag}$~\cite{liang2020shot}} 
&76.4{$_{0.7}$} &54.0{$_{1.1}$} 
&73.2{$_{0.5}$} &54.7{$_{1.2}$} 
&72.8{$_{1.2}$} &59.1{$_{2.2}$} 
&75.4{$_{2.0}$} &53.7{$_{1.4}$} 
&87.6{$_{1.1}$} &52.3{$_{3.3}$} 
&74.6{$_{2.1}$} &49.0{$_{1.8}$} 
&76.7 &53.8 \\

{+T3A~\cite{iwasawa2021t3a}} 
&76.2{$_{1.2}$} &55.3{$_{1.2}$} 
&\underline{75.3}{$_{0.4}$} &\underline{56.8}{$_{1.7}$} 
&\underline{75.1}{$_{1.4}$} &58.6{$_{2.3}$} 
&74.6{$_{2.5}$} &52.9{$_{2.9}$} 
&86.5{$_{1.6}$} &47.6{$_{6.6}$} 
&71.6{$_{3.0}$} &47.1{$_{2.1}$} 
&76.6 &53.1 \\

{+TAST~\cite{jang2022tast}} 
&\underline{77.5}{$_{1.2}$} &\underline{57.1}{$_{1.4}$} 
&74.6{$_{0.9}$} &56.3{$_{1.7}$} 
&71.6{$_{1.2}$} &\underline{59.4}{$_{2.0}$} 
&\underline{77.2}{$_{4.8}$} &53.6{$_{2.9}$} 
&90.2{$_{2.1}$} &\underline{67.6}{$_{4.6}$} 
&\underline{78.2}{$_{4.3}$} &52.3{$_{3.0}$} 
&\underline{78.2} &\underline{57.7} \\

{+UniDG~\cite{zhang2023unidg}} 
&60.1{$_{4.9}$} &43.5{$_{6.1}$} 
&68.7{$_{2.5}$} &46.5{$_{6.3}$} 
&64.1{$_{4.2}$} &35.6{$_{4.9}$} 
&65.3{$_{9.4}$} &43.1{$_{9.5}$} 
&71.2{$_{4.5}$} &21.7{$_{7.4}$} 
&69.3{$_{3.5}$} &44.7{$_{4.4}$} 
&66.5 &39.2 \\

{+DeYO$^{\dag}$~\cite{lee2024deyo}} 
&77.0{$_{1.3}$} &57.0{$_{1.1}$} 
&73.7{$_{0.3}$} &55.5{$_{0.5}$} 
&72.4{$_{2.3}$} &\textbf{60.8}{$_{1.5}$} 
&76.2{$_{2.4}$} &\underline{55.8}{$_{4.3}$} 
&89.9{$_{1.2}$} &61.2{$_{3.0}$} 
&76.9{$_{2.0}$} &\textbf{53.2}{$_{1.5}$} 
&77.7 &57.3 \\

{+TNN~\cite{tnn2024miccai}} &76.9{$_{1.3}$} &57.0{$_{1.5}$} &74.9{$_{0.8}$} &56.5{$_{1.9}$} &74.2{$_{1.3}$} &56.5{$_{2.1}$} &75.6{$_{2.7}$} &54.3{$_{2.5}$} &88.3{$_{1.6}$} &58.1{$_{4.0}$} &73.9{$_{2.2}$} &44.6{$_{2.0}$} &77.3 &54.5 \\

{+EATA$^{\dag}$~\cite{niu2022eata}} 
&75.9{$_{1.0}$} &53.4{$_{1.1}$} 
&73.4{$_{0.6}$} &54.9{$_{1.5}$} 
&74.4{$_{1.2}$} &52.0{$_{1.9}$} 
&75.8{$_{2.2}$} &53.8{$_{3.7}$} 
&91.0{$_{1.1}$} &63.4{$_{3.3}$} 
&73.9{$_{2.7}$} &45.0{$_{1.8}$} 
&77.4 &53.7 \\

{+SAR$^{\dag}$~\cite{niu2023sar}} 
&76.0{$_{1.0}$} &53.4{$_{1.0}$} 
&73.2{$_{0.8}$} &54.6{$_{1.7}$} 
&73.8{$_{1.3}$} &53.7{$_{2.1}$} 
&76.4{$_{2.2}$} &54.0{$_{3.8}$} 
&\underline{91.1}{$_{1.2}$} &65.0{$_{3.3}$} 
&74.5{$_{2.5}$} &48.2{$_{1.0}$} 
&77.5 &54.8 \\

{+FunOTTA} 
&\textbf{78.1}{$_{0.9}$} &\textbf{57.5}{$_{1.2}$} 
&\textbf{75.5}{$_{0.8}$} &\textbf{57.3}{$_{0.3}$} 
&\textbf{75.5}{$_{1.4}$} &55.3{$_{2.8}$} 
&\textbf{77.6}{$_{2.7}$} &\textbf{57.1}{$_{3.7}$} 
&\textbf{91.8}{$_{1.8}$} &\textbf{71.2}{$_{4.4}$} 
&\textbf{78.7}{$_{2.3}$} &\underline{52.8}{$_{2.8}$} 
&\textbf{79.5}{$^{*}$} &\textbf{58.5}{$^{*}$} \\

\midrule
{Upper bound} &80.0{$_{1.1}$} &60.0{$_{1.7}$} &86.4{$_{0.9}$} &68.6{$_{1.5}$} &81.4{$_{1.4}$} &62.9{$_{2.7}$} &90.4{$_{2.9}$} &73.8{$_{3.6}$} &95.3{$_{1.2}$} &75.4{$_{3.5}$} &84.4{$_{2.0}$} &59.4{$_{2.9}$} &84.3 &66.7 \\

\midrule
{ResNet50} 
&74.8{$_{0.9}$} &53.8{$_{1.8}$} 
&74.1{$_{1.4}$} &53.6{$_{1.8}$} 
&74.6{$_{1.6}$} &53.0{$_{6.3}$} 
&73.5{$_{1.3}$} &51.6{$_{2.4}$} 
&87.1{$_{1.6}$} &63.6{$_{4.2}$} 
&77.5{$_{3.2}$} &51.9{$_{4.2}$} 
&76.9 &54.6 \\

{+TENT$^{\dag}$~\cite{wang2020tent}} 
&74.8{$_{1.1}$} &53.2{$_{1.4}$} 
&74.9{$_{1.5}$} &53.9{$_{1.7}$} 
&74.7{$_{1.7}$} &52.0{$_{6.3}$} 
&73.6{$_{1.4}$} &50.5{$_{2.0}$} 
&87.4{$_{1.6}$} &63.1{$_{4.7}$} 
&77.1{$_{3.9}$} &51.9{$_{2.0}$} 
&77.0 &54.1 \\

{+PL$^{\dag}$~\cite{lee2013pseudo}} 
&74.8{$_{0.9}$} &53.6{$_{1.6}$} 
&75.0{$_{1.6}$} &54.2{$_{2.1}$} 
&74.4{$_{1.7}$} &52.7{$_{6.3}$} 
&73.6{$_{1.4}$} &50.5{$_{2.4}$} 
&87.0{$_{1.6}$} &61.6{$_{4.7}$} 
&77.2{$_{3.1}$} &53.2{$_{2.8}$} 
&77.0 &54.3 \\

{+SHOT$^{\dag}$~\cite{liang2020shot}} 
&75.3{$_{0.8}$} &56.6{$_{1.4}$} 
&74.3{$_{1.3}$} &53.7{$_{1.3}$} 
&74.7{$_{1.8}$} &\underline{61.3}{$_{2.6}$} &73.5{$_{1.3}$} &53.5{$_{1.7}$} &86.9{$_{1.3}$} &62.8{$_{2.1}$} &77.7{$_{3.1}$} &53.9{$_{4.6}$} &77.1 &57.0 \\

{+T3A~\cite{iwasawa2021t3a}} 
&\underline{76.2}{$_{1.0}$} &58.0{$_{1.8}$} &75.0{$_{1.4}$} &\underline{55.4}{$_{1.7}$} &\underline{75.0}{$_{1.3}$} &55.5{$_{2.3}$} &\underline{73.9}{$_{1.9}$} &53.0{$_{1.2}$} 
&\textbf{88.7}{$_{1.2}$} &64.9{$_{4.1}$} &77.3{$_{3.7}$} &53.7{$_{4.9}$} &\underline{77.7} &56.8 \\

{+TAST~\cite{jang2022tast}} 
&74.6{$_{1.2}$} &\underline{58.1}{$_{1.6}$} 
&\textbf{75.7}{$_{1.8}$} &54.8{$_{1.7}$} 
&73.4{$_{1.4}$} &57.2{$_{2.6}$} 
&73.8{$_{1.6}$} &52.3{$_{2.0}$} 
&87.6{$_{2.6}$} &\underline{66.1}{$_{5.4}$} 
&78.4{$_{3.0}$} &\underline{54.2}{$_{3.2}$} 
&77.2 &\underline{57.1} \\

{+UniDG~\cite{zhang2023unidg}} 
&60.8{$_{2.6}$} &44.1{$_{2.9}$} 
&67.0{$_{1.2}$} &48.7{$_{3.1}$} 
&64.3{$_{6.1}$} &55.1{$_{2.2}$} 
&66.3{$_{3.5}$} &40.8{$_{0.8}$} 
&64.9{$_{3.5}$} &23.4{$_{4.0}$} 
&70.7{$_{1.4}$} &46.5{$_{2.6}$} 
&65.7 &43.1 \\

{+DeYO$^{\dag}$~\cite{lee2024deyo}} 
&74.5{$_{0.9}$} &54.6{$_{1.4}$} 
&74.1{$_{1.5}$} &53.0{$_{1.6}$} 
&72.8{$_{1.9}$} &60.8{$_{2.9}$} 
&73.7{$_{1.5}$} &\underline{54.7}{$_{1.6}$} 
&\underline{88.5}{$_{1.9}$} &62.1{$_{7.4}$} 
&\underline{79.1}{$_{2.5}$} &\textbf{54.7}{$_{3.3}$} 
&77.1 &56.6 \\

{+TNN~\cite{tnn2024miccai}} &75.3{$_{1.1}$} &57.6{$_{1.5}$} &74.9{$_{1.5}$} &54.8{$_{1.8}$} &74.7{$_{1.5}$} &56.2{$_{2.3}$} &73.6{$_{1.9}$} &53.1{$_{2.5}$} &87.7{$_{1.3}$} &65.8{$_{2.1}$} &78.0{$_{2.3}$} &53.9{$_{3.7}$} &77.2 &56.9\\

{+EATA$^{\dag}$~\cite{niu2022eata}} 
&75.5{$_{0.9}$} &55.0{$_{1.7}$} 
&74.5{$_{1.5}$} &55.0{$_{1.7}$} 
&74.7{$_{1.7}$} &56.3{$_{6.3}$} 
&73.2{$_{1.4}$} &51.8{$_{2.5}$} 
&87.9{$_{1.6}$} &65.6{$_{4.2}$} 
&77.4{$_{1.6}$} &53.0{$_{4.2}$} 
&77.2 &56.1 \\

{+SAR$^{\dag}$~\cite{niu2023sar}} 
&75.5{$_{1.1}$} &55.3{$_{1.8}$} 
&74.0{$_{1.6}$} &54.3{$_{2.0}$} 
&74.9{$_{1.9}$} &56.6{$_{5.9}$} 
&73.2{$_{1.3}$} &52.3{$_{2.0}$} 
&87.8{$_{1.9}$} &65.1{$_{4.5}$} 
&78.2{$_{2.5}$} &53.4{$_{4.0}$} 
&77.3 &56.2 \\

{+FunOTTA} 
&\textbf{76.6}{$_{1.5}$} &\textbf{58.7}{$_{1.5}$} 
&\underline{75.5}{$_{1.4}$} &\textbf{55.9}{$_{1.4}$} 
&\textbf{75.4}{$_{1.0}$} &\textbf{61.6}{$_{2.7}$} 
&\textbf{75.0}{$_{1.6}$} &\textbf{54.8}{$_{1.7}$} 
&87.9{$_{2.2}$} &\textbf{68.0}{$_{4.0}$} 
&\textbf{79.4}{$_{2.4}$} &53.0{$_{4.1}$} 
&\textbf{78.3}{$^{*}$} &\textbf{58.7}{$^{**}$} \\

\midrule
{Upper bound} &79.0{$_{1.2}$} &59.5{$_{2.0}$} &86.5{$_{1.3}$} &66.8{$_{1.6}$} &81.8{$_{1.7}$} &67.8{$_{3.9}$} &89.8{$_{1.8}$} &69.4{$_{2.2}$} &97.8{$_{1.7}$} &76.6{$_{3.1}$} &83.8{$_{3.2}$} &57.2{$_{3.9}$} &84.9 &66.7\\

\bottomrule
\end{tabular}
}
\end{table*}

%% file: ablation_tab.tex
\begin{table}[ht]
    \centering
    \caption{Ablation analysis of the FunOTTA on diabetic retinopathy datasets. The results are averaged across five runs, using ResNet50 as the backbone network.}
    \label{tab:ablation_dr}
    \resizebox{\linewidth}{!}{
    \setlength{\tabcolsep}{2pt}
    \begin{tabular}{@{}cc|c|c|c|c|c|c@{}}
    
    \toprule
         \multirow{2}{*}{$\mathcal{L}_{CCL}$}&\multirow{2}{*}{$\mathcal{L}_{DAL}$}&Messidor-2&ATPOS&IDRID&OIA-DDR&DeepDRiD&Average\\
         
         &&AUC/F1&AUC/F1&AUC/F1&AUC/F1&AUC/F1&AUC/F1\\
        \midrule
        \XSolidBrush&\XSolidBrush&76.5{$_{1.5}$}/54.8{$_{1.8}$}&76.2{$_{1.9}$}/53.6{$_{2.7}$}&71.6{$_{1.0}$}/32.6{$_{0.9}$}&76.2{$_{0.9}$}/39.5{$_{4.1}$}&74.3{$_{1.0}$}/43.0{$_{2.6}$}&75.0/44.7\\
        \Checkmark&\XSolidBrush&76.8{$_{3.0}$}/\textbf{56.6}{$_{2.1}$}&77.5{$_{3.2}$}/55.9{$_{3.3}$}&72.7{$_{1.2}$}/34.7{$_{1.4}$}&72.5{$_{1.0}$}/39.1{$_{4.7}$}&\textbf{77.7}{$_{1.2}$}/44.3{$_{3.5}$}&75.4/46.1\\
        \XSolidBrush&\Checkmark&77.7{$_{1.3}$}/54.1{$_{1.4}$}&80.0{$_{2.1}$}/59.1{$_{2.5}$}&76.2{$_{2.0}$}/37.9{$_{2.2}$}&77.0{$_{1.3}$}/45.7{$_{4.6}$}&77.1{$_{1.0}$}/44.9{$_{2.5}$}&77.6/48.3\\
        \Checkmark&\Checkmark&\textbf{79.1}{$_{2.8}$}/56.3{$_{1.6}$}&\textbf{81.9}{$_{3.0}$}/\textbf{60.0}{$_{2.8}$}&\textbf{77.6}{$_{1.9}$}/\textbf{40.1}{$_{2.2}$}&\textbf{79.0}{$_{1.0}$}/\textbf{51.8}{$_{6.0}$}&77.2{$_{0.9}$}/\textbf{48.4}{$_{3.3}$}&\textbf{78.3}/\textbf{51.3}\\

    \bottomrule
    \end{tabular}
    }
\end{table}

\begin{table}[htbp]
    \centering
    \caption{Ablation analysis of the FunOTTA on glaucoma datasets. The results are averaged across five runs, using ResNet50 as the backbone network.}
    \label{tab:ablation_glaucoma}
    \resizebox{\linewidth}{!}{
    \setlength{\tabcolsep}{2pt}
    \begin{tabular}{@{}cc|c|c|c|c|c|c|c@{}}
    
    \toprule
         \multirow{2}{*}{$\mathcal{L}_{CCL}$}&\multirow{2}{*}{$\mathcal{L}_{DAL}$}&ORIGA&BEH&FIVES&sjchoi86&REFUGE1&PAPILA&Average\\
         
         &&AUC/F1&AUC/F1&AUC/F1&AUC/F1&AUC/F1&AUC/F1&AUC/F1\\
        \midrule
        
        \XSolidBrush&\XSolidBrush&
        75.1{$_{1.8}$}/58.4{$_{1.5}$}&
        74.1{$_{2.2}$}/54.9{$_{1.3}$}&
        73.5{$_{1.0}$}/58.7{$_{2.6}$}&
        75.4{$_{1.6}$}/\textbf{55.4}{$_{1.9}$}&
        88.4{$_{2.3}$}/64.1{$_{3.5}$}&
        71.4{$_{2.5}$}/44.9{$_{3.8}$}&
        76.3/56.1\\
        
        \Checkmark&\XSolidBrush&
        76.1{$_{1.3}$}/57.3{$_{1.7}$}&
        74.2{$_{2.4}$}/55.8{$_{1.5}$}&
        71.9{$_{1.1}$}/57.9{$_{2.0}$}&
        \textbf{78.2}{$_{2.2}$}/54.8{$_{1.6}$}&
        \textbf{88.5}{$_{1.8}$}/\textbf{69.1}{$_{2.9}$}&
        71.3{$_{3.1}$}/47.0{$_{3.5}$}&
        76.7/57.0\\
        
        \XSolidBrush&\Checkmark&
        74.6{$_{1.4}$}/58.1{$_{1.5}$}&
        74.0{$_{1.7}$}/53.9{$_{2.2}$}&
        74.6{$_{1.2}$}/58.2{$_{2.5}$}&
        74.4{$_{1.5}$}/52.9{$_{1.9}$}&
        87.4{$_{2.4}$}/65.7{$_{3.7}$}&
        72.3{$_{2.8}$}/45.3{$_{2.4}$}&
        76.2/55.7\\

        \Checkmark&\Checkmark&\textbf{76.6}{$_{1.2}$}/\textbf{58.7}{$_{1.5}$}&\textbf{75.5}{$_{1.4}$}/\textbf{55.9}{$_{1.4}$}&\textbf{75.4}{$_{1.0}$}/\textbf{61.6}{$_{2.7}$}&75.0{$_{1.6}$}/54.8{$_{1.7}$}&87.9{$_{2.2}$}/68.0{$_{4.0}$}&\textbf{79.4}{$_{2.4}$}/\textbf{53.0}{$_{4.1}$}&\textbf{78.3}/\textbf{58.7}\\

    \bottomrule
    \end{tabular}
    }
\end{table}

%% file: comp_tab_ls.tex
\begin{table}[th]
\caption{{Comparison of methods across domains in label-shifted dataset (DR-LS). Our results are averaged over five runs. The dagger symbol ($\dag$) denotes that we implemented the method by only fine-tuning the classifier weights, as the original approach performed poorly on fundus images. The highest values are highlighted in \textbf{bold}, and the second values are \underline{underlined}.}}
\label{tab:compare_ls}
\centering
\resizebox{\linewidth}{!}{
\setlength{\tabcolsep}{4pt}
\begin{tabular}{l|cc|cc|cc}
\toprule
\multirow{2}{*}{{Dataset}} & \multicolumn{4}{c|}{{DR-LS}} & \multicolumn{2}{c}{\multirow{2}{*}{{Average}}} \\ 
& \multicolumn{2}{c|}{ResNet18} & \multicolumn{2}{c|}{ResNet50} &  \\ 
\midrule
{Metric}  & {AUC} & {F1} & {AUC} & {F1} & {AUC} & {F1}\\
\midrule
Baseline   & 77.9$_{1.6}$ & 34.7$_{1.7}$ & 76.3$_{2.4}$ & 35.6$_{3.4}$ & 77.1 & 35.2\\
+TENT$^\dag$~\cite{wang2020tent}      & 78.1$_{1.7}$ & 34.8$_{1.8}$ & 76.1$_{2.5}$ & 34.2$_{2.5}$ & 77.1 & 34.5\\
+PL$^\dag$~\cite{lee2013pseudo}         & 78.2$_{1.7}$ & 35.1$_{1.8}$ & 76.1$_{2.5}$ & 34.2$_{2.5}$ & 77.2 & 34.7\\
+SHOT$^\dag$~\cite{liang2020shot}       & \underline{79.2$_{2.2}$} & \underline{47.8$_{3.2}$} & \underline{76.9$_{2.1}$} & 45.3$_{3.2}$ & \underline{78.1} & 46.6\\
+T3A~\cite{iwasawa2021t3a}       & 78.2$_{1.1}$ & 40.5$_{4.7}$ & 75.3$_{2.3}$ & 39.2$_{7.8}$ & 76.8 & 39.9\\
+TAST~\cite{jang2022tast}      & 76.7$_{1.7}$ & 47.6$_{6.1}$ & 73.8$_{2.1}$ & \underline{46.5$_{2.9}$} & 75.3 & \underline{47.1}\\
+UniDG~\cite{zhang2023unidg}     & 71.9$_{0.5}$ & 27.4$_{1.2}$ & 65.6$_{1.1}$ & 28.6$_{5.2}$ & 68.8 & 28.0\\
+DeYO$^\dag$~\cite{lee2024deyo}       & 74.8$_{1.8}$ & 33.8$_{3.6}$ & 72.9$_{0.6}$ & 31.9$_{1.3}$ & 73.9 & 32.9\\
+TNN~\cite{tnn2024miccai}       & 78.4$_{1.6}$ & 43.5$_{5.3}$ & 75.1$_{2.7}$ & 38.0$_{5.6}$ & 76.8 & 40.8\\
+EATA$^\dag$~\cite{niu2022eata}       & 78.2$_{1.8}$ & 36.2$_{2.7}$ & 76.3$_{2.3}$ & 36.7$_{4.4}$ & 77.3 & 36.5\\
+SAR$^\dag$~\cite{niu2023sar}        & 78.1$_{1.5}$ & 35.1$_{1.8}$ & 76.2$_{2.4}$ & 35.3$_{3.3}$ & 77.2 & 35.2\\
+FunOTTA   & \textbf{79.8$_{1.8}$} & \textbf{49.1$_{4.9}$} & \textbf{78.2$_{3.2}$} & \textbf{48.3$_{2.4}$} & \textbf{79.0} & \textbf{48.7}\\
\midrule
Upper bound & 87.8$_{2.1}$ & 64.5$_{3.5}$ & 87.1$_{2.9}$ & 63.5$_{3.7}$ & 87.5 & 64.0\\
\bottomrule
\end{tabular}
}
\end{table}